\documentclass{article} %
\usepackage[final]{corl_2026}

\usepackage[utf8]{inputenc} %
\usepackage[T1]{fontenc}    %
\usepackage{hyperref}       %
\usepackage{url}            %
\usepackage{booktabs}       %
\usepackage{amsfonts}       %
\usepackage{nicefrac}       %
\usepackage{microtype}      %
\usepackage[dvipsnames]{xcolor}         %

\usepackage{xfrac}

\usepackage{booktabs}       %
\usepackage{multirow}
\usepackage{microtype}      %
\usepackage{xcolor}         %
\usepackage{xcolor-material}
\usepackage{graphicx}
\usepackage{caption}
\usepackage{wrapfig}
\usepackage[labelformat=simple]{subcaption}
\usepackage{soul}
\IfFileExists{minted2.sty}{%
  \usepackage[frozencache,cachedir=_minted-main]{minted2}%
}{%
  \usepackage[frozencache,cachedir=_minted-main]{minted}%
}

\usepackage{nicefrac}       %
\usepackage{amsfonts}       %
\usepackage{amsmath, bm}
\usepackage{amssymb}
\usepackage{bbm}
\usepackage{cleveref}
\usepackage{fontawesome}
\usepackage{mathtools}
\usepackage[most]{tcolorbox}      %
\usepackage{enumitem}             %

\usepackage{algorithm}
\usepackage{algpseudocode}

\def\eqref#1{equation~\ref{#1}}

\def\1{\bm{1}}

\def\eps{{\epsilon}}

\def\rvd{{\mathbf{d}}}

\def\rvu{{\mathbf{i}}}

\def\rvu{{\mathbf{u}}}
\def\rvv{{\mathbf{v}}}

\def\rvx{{\mathbf{x}}}

\def\vv{{\bm{v}}}

\def\mD{{\bm{D}}}

\def\mI{{\bm{I}}}

\DeclareMathAlphabet{\mathsfit}{\encodingdefault}{\sfdefault}{m}{sl}
\SetMathAlphabet{\mathsfit}{bold}{\encodingdefault}{\sfdefault}{bx}{n}

\def\gA{{\mathcal{A}}}

\def\gD{{\mathcal{D}}}

\def\gM{{\mathcal{M}}}
\def\gN{{\mathcal{N}}}
\def\gO{{\mathcal{O}}}
\def\gP{{\mathcal{P}}}

\def\gR{{\mathcal{R}}}
\def\gS{{\mathcal{S}}}

\def\sD{{\mathbb{D}}}

\newcommand{\E}{\mathbb{E}}

\newcommand{\R}{\mathbb{R}}

\DeclareMathOperator*{\argmin}{arg\,min}

\def\Ind#1{{\mathbbm{1}_{\left[#1\right]}}}

\newcommand{\decidecolortest}[1]{%
    \colorlet{mylightcolor}{MaterialIndigo50}%
    \colorlet{mycolor}{MaterialIndigo500}%
    \colorlet{mydarkcolor}{MaterialIndigo900}%
    \ifstrequal{#1}{blue}{%
        \colorlet{mylightcolor}{MaterialIndigo50}%
        \colorlet{mycolor}{MaterialIndigo500}%
        \colorlet{mydarkcolor}{MaterialIndigo900}%
        }{}%
    \ifstrequal{#1}{red}{%
        \colorlet{mylightcolor}{MaterialRed50}%
        \colorlet{mycolor}{MaterialRed500}%
        \colorlet{mydarkcolor}{MaterialRed900}%
        }{}%
    \ifstrequal{#1}{green}{%
        \colorlet{mylightcolor}{MaterialGreen50}%
        \colorlet{mycolor}{MaterialGreen500}%
        \colorlet{mydarkcolor}{MaterialGreen900}%
        }{}%
    }
\newlist{boxitemize}{itemize}{2}
\tcbset{boxstyle/.style = {
    enhanced,
    breakable,
    sharp corners,
    boxrule = 1.5pt,
}}
\newtcolorbox{mybox}[2][]{
    boxstyle,
    code = {\decidecolortest{#1}%
        \setlist[boxitemize]{leftmargin = *}
        \setlist[boxitemize, 1]{label = {\textcolor{mycolor}{$\filledsquare$}}}
        \setlist[boxitemize, 2]{label = {\textcolor{mycolor}{$\bullet$}}}
    },
    colback = mylightcolor,
    colframe = mydarkcolor,
    title = {#2}
}

\newtcolorbox[auto counter, number within=section]{todolist}[1][]{%
  colback=gray!5!white, colframe=gray!75!black,
  fonttitle=\bfseries, title=TODO #1,
}

\newtcolorbox[auto counter, number within=section]{outlinelist}{%
  colback=blue!5!white, colframe=blue!40!black,
  fonttitle=\bfseries, title=Outline,
}

\newtcolorbox[auto counter, number within=section]{contributionlist}{%
  colback=blue!5!white, colframe=blue!40!black,
  fonttitle=\bfseries, title=Contributions,
}

\newtcolorbox[auto counter, number within=section]{summarylist}{%
  colback=green!5!white, colframe=green!40!black,
  fonttitle=\bfseries, title=Summary,
}

\newtcolorbox[auto counter, number within=section]{graybox}[1][]{%
  colback=gray!10!white, colframe=gray!20!black, coltitle=gray!10!white,
  fonttitle=\bfseries, title=#1,
}

\tcbuselibrary{minted} %
\crefname{snippet}{snippet}{snippets}
\Crefname{snippet}{Snippet}{Snippets}
\newtcbinputlisting[auto counter]{\inputcbminted}[5][]{ %
  breakable,                            %
  listing engine=minted,                %
  listing file={#2},                    %
  listing only,                         %
  minted language={#3},                 %
  title={#4 \hfill \textsf{\scriptsize [Snippet \thetcbcounter ]}},                           %
  label={#5},                           %
  label type=snippet,                      %
  fonttitle=\bfseries,                  %
  fontupper=\small\ttfamily,            %
  colback=gray!5,                       %
  colframe=gray!20!black,               %
  coltitle=gray!10!white,               %
  #1,                                   %
  minted options={
    breaklines=true,
    breaksymbolleft={},
    breakautoindent=true,
    breakindent=0.1em,
    breakanywhere=true,
  }
}

\usepackage[addedmarkup=uline, defaultcolor=magenta, todonotes={textsize=scriptsize, textwidth=3.5cm}, authormarkuptext=name, commandnameprefix=always, xcolor]{changes} %

\definechangesauthor[name={JD}, color=orange]{jd}

\definechangesauthor[name={AK}, color=blue]{ak}

\definecolor{addedlines}{RGB}{200, 255, 200} %
\newcommand{\stt}[1]{{\small{\texttt{#1}}}}
\newcommand{\ssf}[1]{{\small{\textsf{#1}}}}

\setlist{nolistsep}
\author{
  Arjun Krishna\\
  University of Pennsylvania \\
  \texttt{arjk@seas.upenn.edu}
  \And
  Vincent Pacelli\thanks{work done as a postdoctoral fellow at Georgia Institute of Technology, prior to joining Amazon Robotics.
} \\
  Amazon Robotics \\
  \texttt{vpacelli@amazon.com}
  \And
  Dinesh Jayaraman\\
  University of Pennsylvania \\
  \texttt{dineshj@seas.upenn.edu}
}

\crefname{section}{sec.}{secs.}
\Crefname{section}{Sec.}{Secs.}

\title{LEMCA: LLM-Guided Synthesis of Efficient Mode-Switching Control Architectures}
\newcommand{\myalgo}{LEMCA}
\newcommand{\myalgoexpanded}{{\bf L}LM-Guided synthesis of {\bf E}fficient {\bf M}ode-Switching {\bf C}ontrol {\bf A}rchitectures}

\newcommand{\task}[1]{\texttt{#1}}

\begin{document}
\maketitle
\begin{abstract}
Physical control tasks in the natural world, such as driving or object manipulation, frequently exhibit dramatic variations in sensory and compute complexity over time. %
Correspondingly, a natural resource-efficient choice for robot control is to %
dynamically switch between control modes with varying resource allocations.
However, such ``mode-switching controllers'' (MSCs) have historically required laborious, expert-driven design and synthesis for each new task.
Driven by these design difficulties, modern robotic control architectures often fall back to a wasteful ``monolithic'' one-size-fits-all structure, where resource allocation is permanently anchored to the hardest, most resource-intensive task phases.
  To facilitate the design of performant yet efficient MSCs,
  we propose \myalgoexpanded ~(\myalgo).
  \myalgo{} represents MSC designs as interpretable programs to be iteratively refined in an evolutionary loop.
  To evaluate design fitness, we propose MSC-compatible extensions of automated controller synthesis approaches, such as reinforcement learning in simulation.
  \myalgo{} then leverages the semantic priors, reasoning, and coding capabilities of Large Language Models (LLMs) to iteratively edit controller modes, their corresponding sensory-compute resource allocations, and mode transitions.
  Our experiments across diverse control benchmarks show that \myalgo\ consistently discovers strategies that surpass the Pareto frontier of monolithic designs by reclaiming wasted resources during ``easy'' task phases. %
  \myalgo{} thus presents an automated, low-effort path to synthesize resource-efficient MSC designs.

\end{abstract}

\section{Introduction}
\label{sec:intro}
\begin{wrapfigure}{r}{0.25\textwidth}
  \centering
  \vspace{-0.5\baselineskip}
  \includegraphics[width=\linewidth]{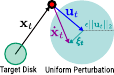}
  \caption{point-to-disk}
  \label{fig:p2d_illustration}
  \vspace{-\baselineskip}
\end{wrapfigure}
We seek a repeatable recipe for synthesizing performant-yet-efficient controllers for new tasks. Consider Erdmann's~\cite{erdmann1995understanding} classical ``point-to-disk'' task, illustrated here. A robot must move the red point from an unknown initial position $x_{t=0}$ on a plane into a target ``disk'' centered on the origin $\mathbf{0}$. When it commands a motion $u_t$, its actual motion vector can deviate from the command within a ball of radius proportional to $\|u_t\|$. Erdmann asks: what is a minimal sensing setup to consistently complete this task? Through an elegant geometric argument, he establishes that a world-centered sensor reporting the full position $x_t$ is a profligate choice. Instead, the robot can complete this task with an ``action-based sensor'' capable of resolving $x_t$'s membership among as few as three equal angular sectors centered at $\mathbf{0}$, each mapping to one task-directed action. Since the time of this work, others~\cite{donald1995information,lavalle2012sensing,mcfassel2020every,majumdar2023fundamental} studying related questions have demonstrated alternative analysis approaches to characterize the sensory requirements of several individual robotic tasks.

Alas, such analyses fall short of our stated goal. For one, most robotic tasks do not permit these elegant analyses, and even when they do, solutions such as ``action-based sensors'' might be impractical in hardware.
Next, sensing is of course only one consideration in resource-efficiency, which should more holistically include computation, energy, latency, etc. Further, we often seek not one solution but a ``Pareto frontier'' family of solutions that optimally trade off task performance for resource costs.
Finally, most real-world tasks like driving and object manipulation naturally have dramatically time-varying resource needs~\citep{krishna2025value,heemels2012introduction}: navigating a cluttered hallway with other dynamic agents may require high-frequency LiDAR and careful path planning, while the same robot traversing a clear hallway could operate nominally with minimal sensing using simple wall-following heuristics. Indeed, we will see in our experiments that even the point-to-disk task has time-varying resource needs.

In practice, we therefore resort to making educated guesses when allocating resources for a robotic control policy. When we have sufficient expertise in a task domain, we might attempt to design ``mode-switching controllers'' (MSCs), that dynamically switch between manually specified task modes with guesstimated design decisions including resource allocations~\citep{nakhaeinia2015hybrid,fierro2001hybrid,burridge1999sequential}. However, even this is out of fashion in general-purpose robotics where per-task laborious expert-driven controller synthesis is prohibitive. Instead, we often fall back to a wasteful ``monolithic'' one-size-fits-all control architecture that implicitly anchors all resource allocations to the hardest, most resource-intensive task phases: our navigation robot above might run high-frequency LiDAR sensing and overly comprehensive path planning computations even in empty hallways.

\begin{figure}[t]
  \includegraphics[width=\linewidth]{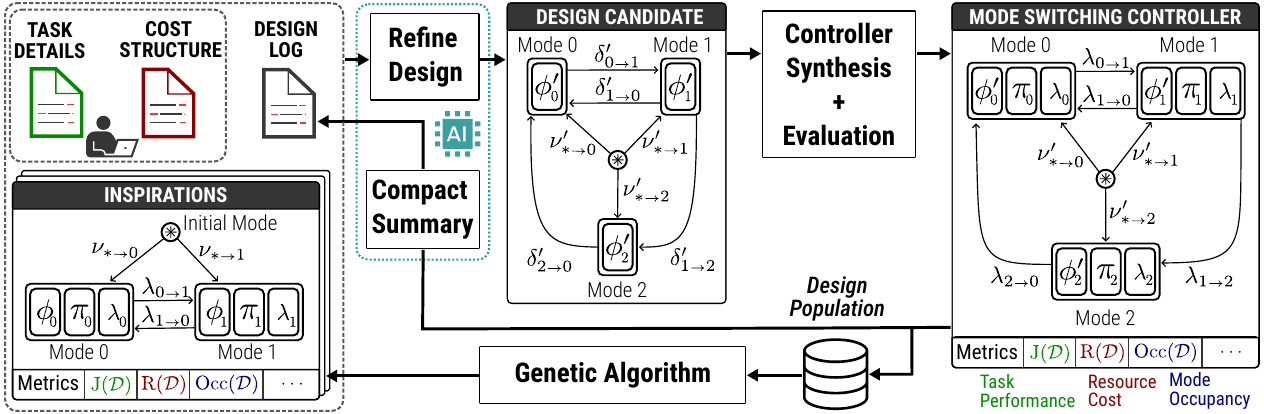}
  \caption{\myalgo\ System Diagram: An LLM-in-the loop evolutionary search pipeline to synthesize and refine mode-switching control architectures represented as a finite state machine specifying various operating modes (containing the sensory-compute configuration $\phi_j$, control policy $\pi_j$, and monitor policy $\lambda_j$), and mode-transitions $\delta$ (see \Cref{sec:method}). A controller-synthesis procedure is used to generate a performant controller adhering to the design specified. The resulting controller's performance metrics serve as feedback for iterative refinement. %
  }
  \label{fig:theme}
\end{figure}

We propose \myalgo, \myalgoexpanded, an automated approach to synthesize MSC designs for new tasks in simulation (see \Cref{fig:theme}). This involves the following key contributions:
\begin{itemize}[leftmargin=*, noitemsep]
    \item \textbf{MSC Design Representation.} We represent MSC designs as programs that specify controller modes, sensory-compute resource allocations, and state-dependent mode transition functions.
    \item \textbf{Problem Formulation as Program Synthesis.} This permits formulating MSC design as a program synthesis problem, opening up new tools for automating MSC design.
    \item \textbf{LLM-Guided Design Sampling.} We propose a task-generic LLM-guided evolutionary search pipeline, \myalgo\, to produce performant-yet-efficient MSC designs. This brings LLM coding and semantic reasoning capabilities to bear on exploring the large MSC design space.
    \item \textbf{MSC-Compatible Reinforcement Learning.} We propose a generic reinforcement learning paradigm to learn controllers subject to an MSC design specification.
    \item \textbf{Empirical Task Suite and Validation.} We design a diverse set of evaluation tasks for MSC design. Besides variations of the point-to-disk task above, this includes rangefinder variants of \texttt{dm\_control} tasks~\citep{tunyasuvunakool2020} and planar navigation.
\end{itemize}
Our experimental results establish that \myalgo\ consistently discovers MSC designs that push beyond the Pareto frontier of monolithic designs.

\section{\myalgoexpanded}
\label{sec:method}
Consider  a task specified by a Markov Decision Process (MDP) $\gM := (\gS, \gA, \gR, \gP, \mu)$, with state space $\gS$, action space $\gA$, reward function $\gR$, transition dynamics $\gP$, and initial state distribution $\mu$.

\paragraph{Sensor Library and Resource Costs}
The agent can observe the environment through a library of sensors $\Phi = \{\varphi_1, \dots, \varphi_n \}$. Here, a ``sensor'' refers broadly to any processed physical measurement, including operations like filtering or feature extraction. Mathematically, each sensor $\varphi_j$ yields an observation in $\gO_j$ distributed according to a history-dependent map $\varphi_j: \gS^* \times \gA^* \to \Delta(\gO_j)$. Activating a subset of sensors $\phi \subseteq \Phi$ incurs a per-timestep resource cost $C(\phi) \in \R^+$ that represents the associated physical overheads such as energy, computation, and communication.

\paragraph{Programmatic Specification of Mode-Switching Controllers~(MSC)}
To enable designers to systematically exploit the non-stationary resource requirements of a task, we adopt a flexible programmatic specification of MSC designs structured as a Finite State Machine~(FSM) $\gD := (\Sigma, \Gamma, \delta, \nu)$:

\begin{center}
    \vspace{-1.1em}
    \noindent
    \begin{minipage}[t]{0.59\textwidth}
        \vspace{0pt}
\begin{itemize}[leftmargin=1em,itemsep=0.7em]
    \item {\bf Control Modes}~($\Sigma$): A discrete set of operating control modes, $\Sigma = \{\sigma^1, \cdots, \sigma^N\}$
    \item {\bf Resource Allocation}~($\Gamma$): Maps control modes to their active sensory-compute configuration, $\Gamma: \Sigma \to 2^\Phi$
    \item {\bf Transition Logic}~($\delta$): Mode transition functions based on an interpretable environment state, $\delta: \Sigma \times \gS \to \Sigma$
    \item {\bf Initial Mode}~($\nu$): Determines initial mode, $\nu: \gS \to \Sigma$
\end{itemize}
    \end{minipage}
    \hfill
    \begin{minipage}[t]{0.39\textwidth}
\begin{minted}{python}
class Design:
  def mode_obs(self) -> dict[int, Config]:
    return {
      0: Config(...),
      1: Config(...),
    }

  def mode_transitions(self)
   -> list[tuple[int, int, callable[[], bool]]:
    def transition_0_1(state) -> bool: ...
    return [(0, 1, transition_0_1)]

  def initial_mode(self, state) -> int:
    return 0
\end{minted}
    \end{minipage}
\end{center}
\paragraph{Design Objective}
\label{sub:design_objective}
The designer's goal is to synthesize a design $\gD$ that minimizes the resource cost $\mathrm{R}(\gD)$, while ensuring the resulting MSC satisfies a minimum performance requirement $\mathrm{J}(\gD) \geq \eps$. \\[0.1em]
Let $\tau = (s_0, \sigma_0, a_0, r_0, \ldots, s_H, \sigma_H, a_H, r_H)$ denote a trajectory of states, modes, actions, and rewards generated by executing the MSC in MDP $\gM$ for $H$ steps. The design objective is:
\begin{align*}
    \argmin_{\gD \in \sD} \mathrm{R}(\gD) &= \E_\tau\left[ \sum_{t=0}^H C(\Gamma(\sigma_t)) \right] \quad\text{s.t. } \mathrm{J}(\gD) = \E_\tau\left[ \sum_{t=0}^H r_t \right]  \geq \eps
\end{align*}
\paragraph{Reinforcement Learning Of Mode-Switching Controllers}
\label{sec:method:synthesis}
To obtain the resource-performance characteristics of a design $\gD$, the designer has to run a synthesis procedure to instantiate an MSC for evaluation.
The MSC $\Pi_\gD$ has to operate on local mode-specific observations $\gO_\sigma$, determined by $\Gamma(\sigma)$, to produce control actions and mode-switching decisions.
Mathematically, $\Pi_\gD$ is a collection of mode-specific controllers $\pi$ and monitors $\lambda$, $\Pi_\gD = \{ (\pi_\sigma, \lambda_\sigma):\gO_\sigma \to \gA \times \Sigma \:|\: \sigma \in \Sigma \}$.
The synthesis procedure should attempt to generate a $\Pi_\gD$ that maximizes the task performance $\mathrm{J}$, while adhering to the sensory-compute constraints specified by the design.
Given the versatility of end-to-end RL, we propose \texttt{MSC-RL} as a generic algorithm to synthesize MSCs for tasks with access to simulators.

\begin{figure}[h]
    \includegraphics[width=\linewidth]{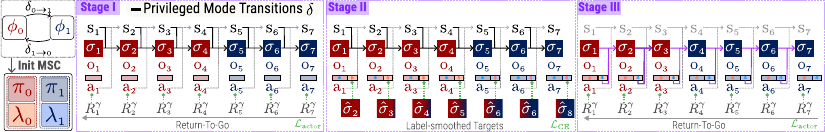}
    \caption{Computation graph of the three stages in MSC-PPO. \stt{\color{blue} [Stage I]} Mode-transitions unroll according to privileged $\delta$ specified by the design $\gD$, corresponding policy networks $\pi_{\sigma_t}$ are activated to sample actions and receive policy gradients. \stt{\color{blue} [Stage II]} Traces from the learned policy are used to train mode-specific monitors $\lambda_{\sigma_t}$ to mimic $\delta$ with temporal label smoothing \stt{\color{blue} [Stage III]} Fine-tunes the policy $\pi$ to adapt to the mode-transitions determined by $\lambda$}
    \label{fig:msc-rl}
\end{figure}

\texttt{MSC-RL} adopts an asymmetric actor-critic paradigm~\citep{pinto2017asymmetric} and instantiates a privileged critic network and a mixture-of-experts style architecture to represent the MSC $\Pi_\gD$ -- where each mode-specific controller $\pi_\sigma$ and monitor $\lambda_\sigma$ is represented by separate networks.
MSC-RL involves three stages of training:
\stt{\color{blue} [Stage I]} Under the privileged mode-transition function $\delta$ specified by the design $\gD$, run any RL algorithm, PPO~\citep{schulman2017proximal} in our case, to optimize the collection of controllers $\{ \pi_\sigma \}$ to maximize $\mathrm{J}(\gD)$.
\stt{\color{blue} [Stage II]} Run the best-performing checkpoint found in \stt{Stage I} with privileged mode-transition function $\delta$ to curate a dataset containing triples: current mode, observation, and next mode $\{ (\sigma_t, o_t, \sigma_{t+1})_{t=0}^H \}$. Train the monitors $\lambda_\sigma$ to mimic the mode transitions with some temporal label smoothing~\citep{muller2019does}, as illustrated in \Cref{fig:msc-rl}.
\stt{\color{blue} [Stage III]} Keeping the monitors $\lambda$ frozen, fine-tune the controllers through RL to adapt to the mode transitions flagged by the learned monitors~\citep{chensequential}.
Across Stages I and III, the MSC controller is frequently evaluated and checkpointed -- the best-performing checkpoint is returned as the representative MSC of the design $\gD$.
As feedback, the designer gets auxiliary performance metrics such as expected resource cost $\mathrm{R}(\gD)$, mode occupancy (\% time spent in each mode), etc., by evaluating this representative MSC.

The controller synthesis procedure is an important part of the design loop, as the assertions on resource-performance tradeoffs are limited by the expressivity and learnability of the chosen controller class.
Additionally, we make choices such as (a) shared neural architecture for controllers/monitors, (b) zero-padding observations to match maximum observation size, for faster training with JAX JIT~\citep{jax2018github} compilation. More implementation details are provided in \Cref{app:msc-rl}

\paragraph{The LLM-Guided Search Procedure} The design space, including modes, sensor configurations, and mode transition functions, is combinatorially large.
Brute force search is infeasible, and so are gradient-based approaches owing to the various discrete choices involved. However, recent advances in LLM-in-loop algorithmic optimization~\citep{novikov2025alphaevolve,lange2025shinkaevolve} offer a viable path to tackle this challenge. These works demonstrate that the semantic priors and reasoning capabilities of modern coding LLMs can be harnessed to serve as effective operators to hypothesize and refine interpretable programs~\citep{zhang2025exploring}, as evidenced in demonstrations of optimizing algorithmic heuristics~\citep{novikov2025alphaevolve,lange2025shinkaevolve}, reward functions~\citep{ma2023eureka}, and curricula~\citep{ma2024dreureka}. We build on these insights to propose \myalgo, \myalgoexpanded, an LLM-guided evolutionary search pipeline to synthesize and refine MSC designs represented as \texttt{Python} programs. The LLM's capabilities are leveraged in two ways as part of the search process: (a) proposing structured edits to synthesize new designs that are likely to improve the objective or gather new insights, and (b) summarizing the insights from evaluations into a compact \texttt{Design Log}, tracking observed trade-offs and open directions, to guide the refinement process -- the prompts for these operations are outlined in \Cref{app:designer:prompts}. The design loop utilizes a genetic algorithm~(GA) with a feasible-first~\citep{deb2000efficient} tournament selection~\citep{goldberg1991comparative,miller1995genetic}; with the crossover handled by an LLM. Due to space limitations, we present the details in \Cref{app:designer:loop}.

\section{Experiments}
\label{sec:results}
We evaluate \myalgo\ across the following tasks with diverse sensory-compute parameterizations:

\task{point-to-disk}: We revisit the task introduced in \Cref{sec:intro}, where a velocity-controlled robot must be steered into a target disk of radius $r^* = 0.01$ as quickly as possible while balancing the costs associated with sensing. The designer can configure sensors that reveal various slices of the state, such as (a) \stt{Cartesian} position, (b) \stt{Radial} distance, and (c) \stt{Sector} presence.
Each sensor type offers distinct tunable noise levels ($\sigma_\text{xy}$, $\sigma_r$, $\sigma_\theta$) and discretizations.
For instance, the designer can specify the number of sectors $n_\text{sec}$ to determine the angular resolution, or
define a sequence of radial boundaries $(d_1, d_2, \dots, d_n)$ to dictate radial distance discretizations.
These configurations naturally introduce different cost-benefit tradeoffs -- for instance, a high-resolution \stt{Sector} detector can help navigate the robot straight into the target disk (minimizing detours), but requires expensive, high-resolution radar scans.
To formalize this, we adopt an information-theoretic cost as a proxy for the per-step resource consumption for each sensor.
We implement a linear-additive cost model that penalizes the information revealed by a sensor configuration $\varphi$, as $C(\varphi) = \alpha I(S; \varphi) + \beta$ (see \Cref{app:p2d} for more details).
To demonstrate that \myalgo\ can adapt to varied cost structures, we evaluate it under two cost specifications: \ssf{CostlyGPS}, where \stt{RadialSector} sensors are substantially cheaper than \stt{Cartesian} sensors, and \ssf{CheapGPS}, where certain \stt{Cartesian} configurations are priced competitively. These setups mirror real-world constraints where sensor modalities carry highly asymmetric costs, often dictated by environmental factors like GPS availability or hardware complexity.

{\task{RF-DMC}}: We consider rangefinder-variants of tasks \task{cartpole-swingup}, \task{cup-catch}, and \task{finger-spin} from the standard \texttt{dm\_control}~\citep{tunyasuvunakool2020,mujoco_playground_2025} suite. Here instead of the state the agent has to operate on $360^\circ$ rangefinder scans from a fixed site -- tip of the pole, center of the cup, and fingertip.
These variants capture many practical characteristics of real systems such as noisy perception, non-linear dynamics, non-linear sensor fusion, etc., while offering a fast-to-simulate setup to study the impacts of sensor noise~\citep{thrun2005probabilistic}, resolution, and history.
The designer is allowed to tune the quality $q$, resolution $n$, and history buffer size $h$ of rangefinder scan for each mode.
The quality refers to discrete operating points that map to distinct noise characteristics, with {\sf A} being the highest grade (minimal noise) and {\sf F} the lowest, high-noise configuration.
Each operating point, with associated SNR, determines the energy consumption of a single rangefinder measurement.
The per-step resource cost of a configuration scales as: $n \cdot \left(\text{Energy}(q) + c_\text{mem}\log h\right)$, see \Cref{app:dmc}.
Task performance is measured by the episodic return normalized by the oracle's episodic return. The returns track the fraction of the episode where the pole is balanced upright, the ball is caught, or the spinner spins above a target velocity -- each condition offers some slack without qualitatively altering the behavior.

{\task{clutter-nav}}:
We consider a 2D planar navigation task in which a disk-shaped robot, governed by unicycle dynamics and equipped with a noisy radar sensor, must navigate to randomly sampled goals while avoiding other static and dynamic disk-shaped obstacles.
The onboard radar sensor detects obstacles within its line of sight and reports their positions and approach velocities subject to measurement noise.
Control synthesis is driven by a Model Predictive Path Integral (MPPI)~\citep{williams2015model} planner operating under a certainty-equivalence assumption, which treats the noisy radar scans as the true state of the environment.
Additionally, the mode-transition functions or monitors specified in the design operate directly over these noisy radar scans -- alleviating the need for any learning.
The design space allows the joint configuration of both {\small \sf Sensor} and {\small \sf Planner} parameters.
For the {\small \sf Sensor}, the configurable parameters include: reliable detection range $R$, the maximum range $r_{\max}$, and the field of view $\vartheta \in [\pi/3, 2\pi]$. The {\small \sf Planner} configuration dictates the number of sampled rollouts $N$,  horizon $H$, refinement iterations $I$, and re-plan period $p$ (where the planner is invoked every $p$ steps to amortize sensing and compute overheads).
Consequently, the per-step resource cost is modeled as a weighted sum of the energy required for reliable sensing ($\propto R^2$), the sensor coverage area ($\propto r_{\max}^2 \cdot \vartheta$), and the computational load ($\propto N \cdot H \cdot I / p$).
Finally, the task performance is quantified as the {\it reciprocal} of the 90th percentile collision rate across 1000 evaluation episodes, subject to a minimum success rate of $80\%$; see \Cref{app:nav} for more details.

\begin{figure}[h]
    \centering
    \includegraphics[width=\linewidth]{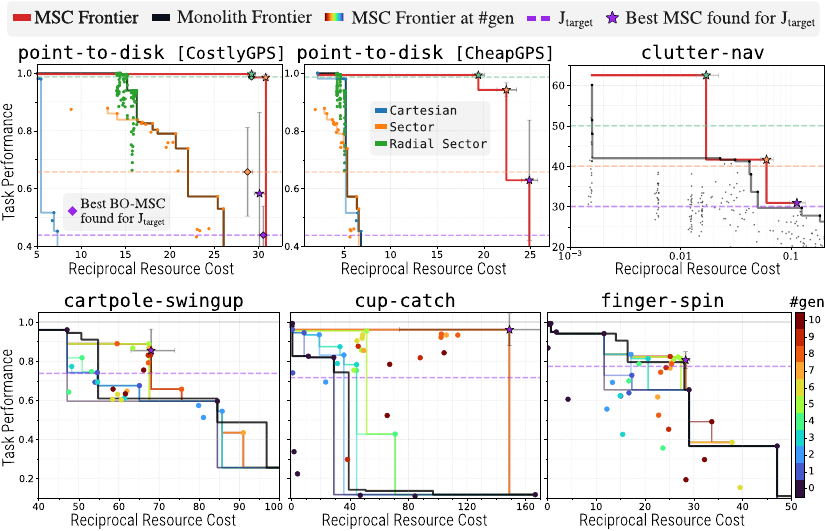}
    \caption{Performance-Resource curve for the evaluated tasks. \myalgo\ consistently identifies MSC designs that dominate the Monolith frontier. The errorbars depict $\pm1\sigma$ bands of 1000 rollouts.}
    \label{fig:pareto}
\end{figure}

\paragraph{\myalgo\ discovers cost-effective MSCs} Across tasks, we find that \myalgo\ consistently identifies MSC designs that push the frontier spanned by monolithic designs obtained by a dense parameter sweep~(\Cref{fig:pareto}).
In most settings, \myalgo\ discovers MSCs that reduce the resource costs by $\geq 20\%$ compared to suitable cost-effective monolithic designs. Notably, in the \task{clutter-nav} task, we observe 10x reductions in resource costs, highlighting the potential of MSC controllers to exploit the non-stationary resource requirements of tasks and the effectiveness of \myalgo\ in synthesizing such strategies. For the \task{RF-DMC} tasks, we additionally visualize the trajectory of designs synthesized by \myalgo\ on the performance-resource plane in \Cref{fig:pareto} (bottom row). This illustrates that \myalgo\ consistently expands the Pareto frontier of MSC designs over generations in the direction that pushes for resource efficiency subject to performance constraint. For \task{point-to-disk}\ssf{[CostlyGPS]}, we additionally contrast the best designs found by \myalgo\ with those found by running BayesOpt~\citep{olson2025ax} on parameterized templates that incorporate structures found in LLM-synthesized designs. While the best designs found by BayesOpt~(BO-MSC) show marginal gains in resource costs, arriving at those designs takes more evaluations ($\approx 100$) and requires pre-specified templates~(see \Cref{app:compare:bo}).

\begin{figure}[ht]
    \includegraphics[width=\textwidth]{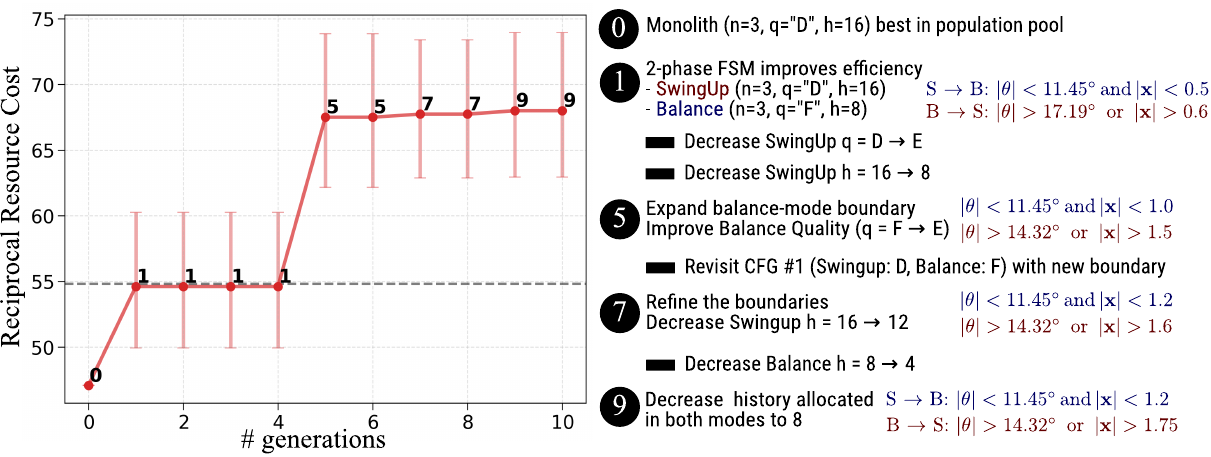}
    \caption{Design decisions and refinements by \myalgo\ over generations for \task{cartpole-swingup}.}
    \label{fig:trace}
\end{figure}

\paragraph{\myalgo\ discovers cost-adaptive designs}
\begin{figure}
    \centering
    \includegraphics[width=\linewidth]{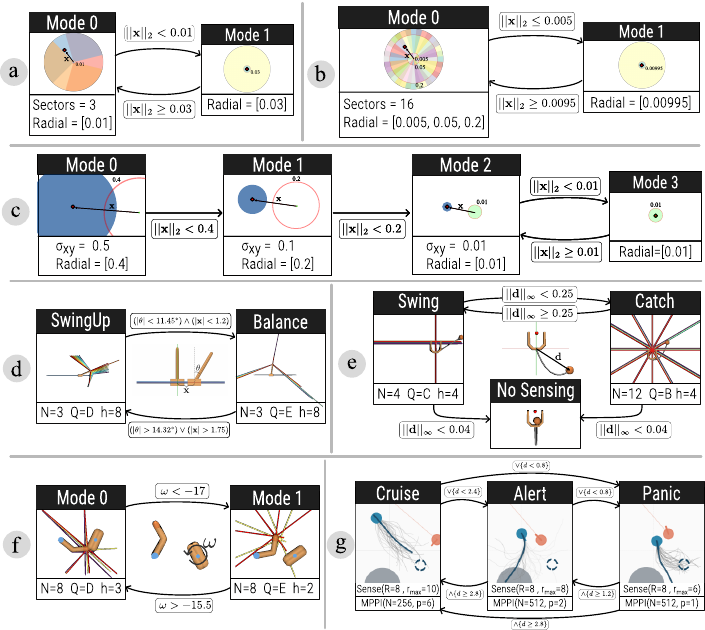}
    \caption{Salient MSC designs identified by \myalgo\ for different tasks. More designs and rollout videos are available at \url{https://lemca-robotics.github.io/}.}
    \label{fig:designs}
\end{figure}
Across the evaluated tasks, \myalgo\ consistently synthesizes highly performant and effective MSC designs by exploiting task and cost structures.
For \task{point-to-disk}, it recognizes that sensing is entirely inconsequential once the  robot enters the target disk.
Under the \ssf{CostlyGPS} cost structure, \myalgo\ adopts a two-mode strategy using \stt{RadialSector} sensors~(\Cref{fig:designs}-(a,b)) for $\mathrm{J}_\text{target}=0.44 \text{ and } 0.98$ -- partitioning the state space into a target disk mode and an outer mode. Based on the performance target, \myalgo\ determines suitable \stt{Sector} and \stt{Radial} discretizations to orchestrate a progressive braking strategy.
Conversely, for \task{point-to-disk}\ssf{[CheapGPS]} at $\mathrm{J}_\text{target}=0.66$, the best design (\Cref{fig:designs}-(c)) leverages affordable \stt{Cartesian} sensors along with \stt{Radial} bands to orchestrate a ``foveation'' strategy, deploying more expensive, less noisy \stt{Cartesian} sensors closer to the target.
In \task{cartpole-swingup}, \myalgo\ settles on a dual-mode strategy (\Cref{fig:designs}-(d)) matching the natural ``swingup'' and ``balance'' phases, utilizing a higher-quality rayscan to swing the pole up before transitioning to a noisier rayscan for upright stabilization.
For \task{cup-catch}~(\Cref{fig:designs}-(e)), a three-mode strategy is dictated by the $L_\infty$ distance between the ball and the cup (tethered by a 0.3m rope);
the MSC activates only four low-quality rangefinders when the rope is nearly taut and dynamics are highly predictable,
transitions to a higher-quality, higher-resolution rayscan as the ball swings over the cup for the catch, and shuts off the sensors entirely once successful.
In \task{finger-spin}~(\Cref{fig:designs}-(f)), the system implements a two-mode strategy that starts with the cost-optimal monolith configuration to strike the spinner above the target velocity of 15 rad/s and switches to a cheaper, lower-quality configuration between subsequent strikes.
Finally, for \task{clutter-nav}~(\Cref{fig:designs}-(g)), a three-mode configuration exploits the natural task affordances: a ``cruise'' mode uses cheaper \ssf{Planner} settings and slower re-planning when no obstacles are in sight, an ``alert'' mode switches to active steering when an obstacle nears, and a high-compute ``panic'' mode maximizes reactivity if the agent becomes dangerously close to an obstacle. These just highlight a few salient designs; more details and rollout visualizations can be found on the project website: \url{https://lemca-robotics.github.io/}

\paragraph{\myalgo\ performs effective refinement} In \Cref{fig:trace}, we visualize the sequential design decisions made by \myalgo\ throughout the refinement process for the \task{cartpole-swingup} task. \myalgo\ begins by establishing reasonable initial choices for both the mode configurations and boundaries. It subsequently experiments with adjustments to various configuration parameters—namely quality, history, and mode boundaries—ultimately arriving at the final configuration illustrated in \Cref{fig:designs}(d).
This execution trace demonstrates that modern reasoning and code-generation LLMs (\stt{gemini-3.1-pro} in our experiments) can effectively navigate the design space to optimize for resource efficiency.
Although we implement a GA + Design Log framework for \myalgo, many other agentic loops merit future exploration. Our ablation study of alternative LLM-in-the-loop variants (\Cref{app:compare:llm}) demonstrates that the GA framework is essential, while the influence of the design log is inconclusive.

\section{Related Work}
\paragraph{Resource-Efficient Hybrid, Event- and Self-Triggered Control.}
Hybrid control and event-triggered control provide a framework to optimize resource consumption of controllers by exploiting task-specific dynamics. For example, hybrid impulse systems restrict actuation to critical instants to conserve propellant \citep{wibben2016terminal}, alternate between planning and reactive policies for navigation \citep{nakhaeinia2015hybrid}, or switch sensory modalities based on contact state \citep{zhao2025touch}. Similarly, networked control methods—including Lebesgue sampling \citep{astrom2002comparison} and event- or self-triggered scheduling \citep{tabuada2007event, anta2010sample, heemels2012introduction}—reduce communication bandwidth by restricting feedback to threshold-crossing events.
A fundamental limitation across these frameworks is that the event specifications, available modes, and resource strategies are hand-crafted by experts. Recent works employing learning within these frameworks \citep{krishna2025value, baumann2018deep, sun2022deep, treven2024sense} typically optimize isolated parameters like trigger thresholds or sampling rates. We propose a more holistic, automated pipeline to jointly synthesize control modes, their transitions, and resource allocations that are tailored to the task performance-cost specifications.

\paragraph{Hierarchical RL.} Task decomposition~\citep{sutton1999between,stolle2002learning} and behavior chaining~\citep{burridge1999sequential,bagaria2019option,chensequential,marzari2021towards} traditionally target long-horizon policy learning and skill re-usability.
However, these hierarchical frameworks typically ignore the operational overhead of the behaviors themselves, and default to monolithic configurations across skills. In contrast, we use the MSC specification to iteratively optimize a controller's inference-time sensory-compute resources.

\paragraph{LLMs for Program Synthesis.}
A growing literature uses code-generating LLMs as operators inside evolutionary search. Eureka~\citep{ma2023eureka} evolves reward code to exceed expert reward engineering. FunSearch~\citep{fawzi2023funsearch}, AlphaEvolve~\citep{novikov2025alphaevolve}, and ShinkaEvolve~\citep{lange2025shinkaevolve} generalize this to the discovery of algorithmic artifacts. \myalgo\ inherits the LLM-as-evolutionary-operator methodology from this lineage but differs in its target: \myalgo\ synthesizes MSC programs under a constrained optimization setting, aiming to minimize resource consumption while retaining functionality. %

\section{Discussion}
By framing the design of MSCs as a program synthesis problem, \myalgo\ demonstrates how LLM-guided evolutionary search can automate the synthesis of efficient-yet-performant MSCs across tasks with different dynamics, sensory modalities, cost structures, and synthesis procedures.
\paragraph{Limitations} This first iteration of our framework makes a few simplifying assumptions: (a) the monitor and controllers in a mode share the same sensor suite, (b) the cost of synthesizing actions and switching decisions is homogeneous, and (c) mode-switching costs are negligible. While these assumptions limit the scope of designs that can be explored, it is straightforward to extend \myalgo{} to decouple configurations and factor the non-homogeneous cost characteristics.
Next, our resource-performance characterizations are fundamentally bounded by the synthesis procedure. The expressivity and learnability of the synthesis procedure, such as \texttt{MSC-PPO}, determine what tradeoffs are realizable -- an unreliable synthesis can mislead the search.
Additionally, with \texttt{MSC-PPO} each evaluation requires a full training run, making the pipeline computationally expensive when simulation is costly -- \task{RF-DMC} tasks require $\approx 0.5$ L40 GPU hours per evaluation. Finally, our ablations of the designer loop suggest that while the design log generally aids refinements, it can, in some settings, limit exploration and result in suboptimal choices -- suggesting more principled approaches to context management might be needed depending on the problem.

\paragraph{Future Directions} While the proposed pipeline primarily focuses on resource allocation as a discrete optimization problem handled by an LLM, a promising avenue for future work is integrating continuous optimization tools -- such as BayesOpt or gradient-based optimization for templated design parameterizations -- to alleviate the LLM's computational burden. Additionally, incorporating more efficient and reliable controller synthesis algorithms~\citep{hu2024privileged,hussing2026behavior} represents an important step in scaling this pipeline to a broader range of tasks.
\clearpage
\newpage
\section*{Acknowledgment}
This research was supported by DARPA TIAMAT HR0011249042, NSF CAREER 2239301, ONR N00014-22-1-2677, and NSF SLES 2331783.

\bibliography{references}

\appendix
\section*{Table of Contents}
\vspace{-0.5em}\Cref{app:impl}: Implementation Details \\[0.1em]
\hspace*{4em} \labelcref{app:msc-rl}: \texttt{MSC-PPO}: Overview and Hyperparameters \hfill (page \pageref{app:msc-rl}) \\[0.1em]
\hspace*{4em} \labelcref{app:designer:prompts}: Prompts used in \myalgo{} \hfill (page \pageref{app:designer:prompts}) \\[0.1em]
\hspace*{4em} \labelcref{app:designer:loop}: \myalgo{}: Overview and Hyperparameters \hfill (page \pageref{app:designer:loop}) \\[0.5em]
\ssf{Task Details} \\[0.1em]
\hspace*{1em}\Cref{app:p2d}: \task{point-to-disk} \hfill (page \pageref{app:p2d}) \\[0.1em]
\hspace*{1em}\Cref{app:dmc}: \task{RF-DMC} (rangefinder-variants of \texttt{dm\_control} tasks)\hfill (page \pageref{app:dmc}) \\[0.1em]
\hspace*{1em}\Cref{app:nav}: \task{clutter-nav} \hfill (page \pageref{app:nav}) \\[0.5em]
\Cref{app:compare}: Points of comparison \\[0.1em]
\hspace*{4em} \labelcref{app:compare:bo}: \texttt{BO-MSC}: BayesOpt on \task{point-to-disk}\ssf{[CostlyGPS]} \hfill (page \pageref{app:compare:bo}) \\[0.1em]
\hspace*{4em} \labelcref{app:compare:llm}: Alternative LLM-in-the-loop mechanisms \hfill (page \pageref{app:compare:llm}) \\[0.5em]
\Cref{app:related_work}: Extended Related Work \hfill (page \pageref{app:related_work}) \\

\begin{mybox}[green]{\sffamily\small Project Website}
\textbf{Link:} \url{https://lemca-robotics.github.io/} \\[0.1em]
The website contains interactive media of \myalgo{} generated designs and its rollouts.
\end{mybox}

\section{Implementation Details}
\label{app:impl}
\subsection{\texttt{MSC-PPO}}
\label{app:msc-rl}
We use an end-to-end JAX-JIT compiled PPO~\citep{schulman2017proximal} implementation as the underlying RL trainer-loop for synthesizing MSCs in tasks \task{point-to-disk} and \task{RF-DMC}.
The implementation is based on open source libraries: \texttt{JAX}~\citep{jax2018github}, \texttt{Optax}~\citep{deepmind2020jax}, \texttt{Equinox}~\citep{kidger2021equinox}, and \texttt{mujoco-playground}~\citep{mujoco_playground_2025}.
The high-level overview of the MSC-PPO trainer is described in \Cref{alg:msc-ppo,alg:label-smooth} and the hyperparameters are outlined in \Cref{tab:msc-ppo:hyperparameters}.
The code will be made available on the website upon publication.

\paragraph{Architecture} \texttt{MSC-PPO} implements an asymmetric actor-critic paradigm~\citep{pinto2017asymmetric}.
The MSC $\Pi_\gD$ is represented as a collection of actor $\pi_\sigma$ and monitor $\lambda_\sigma$ networks, each sharing the same architecture to facilitate JIT compilation.
Similarly, the mode-specific observations are zero-padded to a common~(maximum) dimension to fix shapes for JIT compilation, and a fixed-length FIFO buffer supplies sensory readings for the requested mode-specific $h$ history length.
A single \emph{privileged} critic $V$ operates on the full environment state $s_t$ and is shared across modes and represents two-hot encoded value targets over a fixed symlog support~\citep{hafner2023mastering}.
The actors ingest mode-specific observations to represent a $\tanh$-squashed action distribution with state-dependent variance represented as a diagonal Gaussian -- all tasks have a bounded action space $[-1, 1]^{|\gA|}$.
The monitors operate on mode-specific observations and executed action to produce logits over the next $N$ modes.
The networks are MLPs that use LayerNorm post-activations -- the depth and width are configured per task~(\Cref{tab:msc-ppo:hyperparameters}). The networks operate on normalized observations with per-mode normalization statistics tracked online over rollouts.

\paragraph{Training} The three stages of \texttt{MSC-PPO}~(\Cref{sec:method:synthesis}) share a common rollout-and-update structure (\Cref{fig:msc-rl}, \Cref{alg:msc-ppo}).
Rollouts collect $S$ steps across $E$ parallel environments and compute advantages with GAE~\citep{schulman2015high}.
We adopt a phasic, decoupled update~\citep{cobbe2021phasic}: where for every batch ($E \times S$) the critic is updated for $N_V$ epochs and the (active) actors for $N_\pi$ epochs, each epoch involves running gradient descent over randomly permuted and grouped minibatches of size $M$.

\stt{\color{blue}[Stage I]} trains $\{\pi_\sigma\},V$ under the privileged transition $\delta$.
\stt{\color{blue}[Stage II]} freezes the policy, transitions modes according to $\delta$, and trains monitors $\{\lambda_\sigma\}$ to predict the next mode by minimizing cross-entropy against temporally label-smoothed targets~\citep{muller2019does} for $N_\lambda$ epochs.
The smoothing (\Cref{alg:label-smooth}) convolves the one-hot next-mode sequence with a 1-D Gaussian kernel of width 7 along time, and renormalizes the categorical distribution over modes for each step.
This blends mode labels within a $\pm 3$-step window around each transition so that the monitor can learn softer decision boundaries near impending mode switches.
\stt{\color{blue}[Stage III]} freezes the monitors and fine-tunes $\{\pi_\sigma\},V$ with the learned transitions $\sigma_{t+1}=\arg\max_\sigma \lambda_{\sigma_t}(o_t,a_t)$.

Networks are optimized with AdamW (Nesterov Momentum), with the weight decay applied only to the weight matrices.
The biases, normalization scales, and policy log-std use vanilla Adam.
The optimizer is configured to use global gradient-norm clipping with learning-rate warmup schedules~(\Cref{tab:msc-ppo:hyperparameters}) that are restarted every stage.
Throughout Stages I and III the controller is periodically evaluated and the checkpoint with the best episodic return is carried forward to the next stage; the final best checkpoint is returned as the representative MSC $\Pi_\gD$ for the training run.
When running MSC-PPO with multiple seeds, we pick the best performing checkpoint (based on episodic return) across seeds as the representative MSC for that design.

\begin{algorithm}[H]
\caption{\small \texttt{MSC-PPO}: Controller synthesis for a design $\gD$}
\label{alg:msc-ppo}
\small
\textbf{Input:} MDP $\gM$, design $\gD=(\Sigma,\Gamma,\delta,\nu)$, sensor library $\Phi$ \\[0.3em]
\textbf{Networks:} \\
per-mode actors $\{\pi_\sigma: \gO_\sigma \to \Delta(\gA)\}$, per-mode monitors $\{\lambda_\sigma: \gO_\sigma \to \Delta(\Sigma)\}$, privileged critic $V: \gS \to \R$ \\[0.3em]
\textbf{Parameters:} \#envs $E$, rollout length $S$, \#batches $N_\mathrm{I},N_\mathrm{II},N_\mathrm{III}$, \#epochs $N_V,\,N_\pi,\,N_\lambda$ \vspace{0.3em}
\begin{algorithmic}[1]
    \setlength{\itemsep}{0.15em}
    \Procedure{Rollout}{$\texttt{transition\_fn}$}\Comment{$E{\times}S$ transitions}
        \State $\sigma_0 \gets \nu(s_0)$;\enspace at each $t$: $o_t \sim \Gamma(\sigma_t)$,\, $a_t \sim \pi_{\sigma_t}(\cdot\mid o_t)$,\, \texttt{env.step} $\gM$,\, $V$ on privileged $s_t$
        \State $\sigma_{t+1} \gets \texttt{transition\_fn}(\sigma_t, o_t, a_t, s_{t+1})$
        \State \Return transitions $\left\{ (s_t, r_t, \sigma_t, o_t, a_t) \right\}$
    \EndProcedure
    \vspace{0.3em}
    \State // \stt{\color{blue}{[Stage I]}}{\color{gray}{: train actors under privileged transitions}}
    \For{$k=1$ \textbf{to} $N_\mathrm{I}$}
        \State $\mathcal{B}\gets\Call{Rollout}{\;(\sigma,\_,\_,s') \mapsto \delta(\sigma, s')\;}$;\enspace compute GAE
        \State update $V$ for $N_V$ epochs, then update active actors $\{\pi_{\sigma_t}\}$ for $N_\pi$ epochs (PPO)
        \State periodically evaluate; track best-return checkpoint
    \EndFor;\enspace \textsf{\footnotesize [load best checkpoint]}
    \vspace{0.2em}
    \State // \stt{\color{blue}{[Stage II]}}{\color{gray}{: distill monitors (actors frozen)}}
    \For{$k=1$ \textbf{to} $N_\mathrm{II}$}
    \State $\mathcal{B}\gets\Call{Rollout}{\;(\sigma,\_,\_,s') \mapsto \delta(\sigma, s')\;}$; \enspace Temporal Label Smoothing (\Cref{alg:label-smooth}): $\sigma_{t+1}\!\to\!\hat{\sigma}_{t+1}$ %
        \State update $\{\lambda_\sigma\}$ for $N_\lambda$ epochs with cross-entropy loss to imitate $(o_t,a_t)\!\mapsto\!\hat{\sigma}_{t+1}$
    \EndFor
    \vspace{0.2em}
    \State // \stt{\color{blue}{[Stage III]}}{\color{gray}{: refine actors under learned monitors ($\lambda$ frozen)}}
    \For{$k=1$ \textbf{to} $N_\mathrm{III}$}
        \State $\mathcal{B}\gets\Call{Rollout}{\;(\sigma,o,a,\_) \mapsto \arg\max_{\sigma'}\lambda_\sigma(o,a)\;}$;\enspace compute GAE
        \State update $V$ for $N_V$ epochs, then update active actors for $N_\pi$ epochs; track best-return checkpoint
    \EndFor;\enspace \textsf{\footnotesize [load best checkpoint]}
    \State \Return $\Pi_\gD=\{(\pi_\sigma,\lambda_\sigma)\}$
\end{algorithmic}
\end{algorithm}

\begin{algorithm}[H]
    \caption{\small Temporal Label Smoothing (\stt{\color{blue}{[Stage~II]}} monitor targets)}
    \label{alg:label-smooth}
\small
\textbf{Input:} one-hot next-mode sequence $Y \in \{0,1\}^{T \times N}$ (row $Y_{t,:}=\mathrm{onehot}(\sigma_{t+1})$) \\[0.3em]
\textbf{Parameters:} kernel width $w=7$, scale $f_s=0.9$\vspace{0.3em}
\begin{algorithmic}[1]
    \setlength{\itemsep}{0.15em}
    \State $\tau \gets \mathrm{linspace}(-3,\,3,\,w)$;\enspace $g_i \gets \exp\!\big(-\tfrac{1}{2}\,(\tau_i / f_s)^2\big)$ \Comment{1-D Gaussian Kernel, length $w$}
    \For{each mode $k \in \{1,\dots,N\}$}
        \State $\widetilde{Y}_{:,k} \gets g * Y_{:,k}$ \Comment{1-D convolution along time, \texttt{'same'} padding}
    \EndFor
    \State $\widehat{Y}_{t,:} \gets \widetilde{Y}_{t,:} \,/\, \textstyle\sum_{k} \widetilde{Y}_{t,k}$ \quad $\forall t$ \Comment{renormalize distribution over modes}
    \State \Return smoothed targets $\widehat{Y}$ (each row $\hat{Y}_{t,:}\in\Delta(\Sigma)$) \Comment{cross-entropy targets $\hat{\sigma}_{t+1}$}
\end{algorithmic}
\end{algorithm}

\begin{table}[H]
    \centering
    {\renewcommand{\arraystretch}{1.2}\small
    \begin{tabular}{llcc}
        \multicolumn{2}{l}{\bf Hyperparameter} & {\bf \task{point-to-disk}} & {\bf \task{RF-DMC}} \\ \hline
        \multicolumn{4}{l}{\sffamily\scriptsize\underline{Architecture (actor / critic / monitor MLPs)}} \\
        & depth (\# hidden layers)  & $3$ / $3$ / $3$ & $4$ / $4$ / $4$ \\
        & width (hidden dim) & $64$ / $256$ / $64$ & $512$ / $512$ / $512$ \\
        & activation & \multicolumn{2}{c}{SiLU} \\
        & critic value head & \multicolumn{2}{c}{two-hot, $[-20,20]$, $255$ bins} \\[0.5em]

        \multicolumn{4}{l}{\sffamily\scriptsize\underline{PPO Parameters}} \\[0.5em]
        $E$ & parallel environments & $32,768$ & $8,192$ \\
        $S$ & rollout steps & $16$ & $24$ \\
        $S_d$ & monitor distill rollout steps & $32$ & $32$ \\
        $M$ & minibatch size & $4,096$ & $1,024$ \\
        $N_V/N_\pi/N_\lambda$ & critic / actor / monitor epochs & $3$ / $1$ / $1$ & $5$ / $2$ / $3$ \\
        $\gamma$ & reward discount factor & \multicolumn{2}{c}{$0.99$} \\
        $\lambda$ & GAE parameter & \multicolumn{2}{c}{$0.95$} \\
        $\epsilon_\text{clip}$ & PPO clip ratio & \multicolumn{2}{c}{$0.2$} \\
        $c_\text{ent}$ & entropy coefficient & \multicolumn{2}{c}{$10^{-3}$} \\[0.5em]

        \multicolumn{4}{l}{\sffamily\scriptsize\underline{Stage Budgets}} \\
        $N_\mathrm{I}$ & \stt{\color{blue}{[Stage I]}}\hspace{1em} (\,train\,) & $100$ & $500$ \\
        $N_\mathrm{II}$ & \stt{\color{blue}{[Stage II]}} \hspace{0.2em} (\,distill monitor\,) & $100$ & $200$ \\
        $N_\mathrm{III}$ & \stt{\color{blue}{[Stage III]}} (\,refine\,) & $100$ & $200$ \\[0.5em]

        \multicolumn{4}{l}{\sffamily\scriptsize\underline{Optimization}} \\
        & optimizer & \multicolumn{2}{c}{AdamW (Nesterov Momentum)} \\
        & betas     & \multicolumn{2}{c}{$(\,0.9,\:0.999\,)$} \\
        & learning rate & \multicolumn{2}{c}{$5{\times}10^{-4}$} \\
        & weight decay (weight matrices only) & \multicolumn{2}{c}{$10^{-4}$} \\
        & gradient clip (global norm) & \multicolumn{2}{c}{$10.0$} \\
        & actor schedule (warmup / cooloff steps) & \multicolumn{2}{c}{linear $1000$ / cosine $1000$} \\
        & critic schedule (warmup steps) & \multicolumn{2}{c}{linear $1000$} \\
        & monitor schedule (warmup steps) & \multicolumn{2}{c}{linear $100$} \\[0.5em]

        \multicolumn{4}{l}{\sffamily\scriptsize\underline{Monitor Label Smoothing}} \\
        & Gaussian kernel width & \multicolumn{2}{c}{$7$} \\
        & Gaussian kernel scale $f_s$ & \multicolumn{2}{c}{$0.9$} \\[1em]
    \end{tabular}}
    \caption{\texttt{MSC-PPO} hyperparameters across tasks.}
    \label{tab:msc-ppo:hyperparameters}
\end{table}
\vspace{-2em}
\subsection{Prompts}
\label{app:designer:prompts}
\paragraph{Refine Design} We base our system prompt for design refinement on ShinkaEvolve's~\citep{lange2025shinkaevolve} \texttt{diff} generation template -- where an LLM is prompted to generate targeted code-diffs to the current solution.
The system message for {\bf Refine Design}~(\Cref{app:designer:prompts:system-refine}) includes user provided task \& cost specification, objective, and design template (e.g. \Cref{app:designer:prompts:template}).
Then as a user message, \myalgo{} supplies: a current design $\gD$~(\Cref{app:designer:prompts:user-refine}), list of inspirations~(\Cref{app:designer:prompts:insp-refine}), design log, and corresponding metrics.
The LLM responds with refinements to the current design $\gD$ based on an ``improve''~(\Cref{app:designer:prompts:obj-improve}) or ``explore''~(\Cref{app:designer:prompts:obj-explore}) objective (see \Cref{alg:myalgo}).
The proposed \texttt{diff} is parsed and applied to the current design $\gD$ to generate a new design $\gD_{\text{new}}$.
The new design code undergoes validation -- where all code paths are tested with mock inputs -- to ensure it is well-formed. If validation fails, the LLM is prompted again with an error stack trace to fix the proposed code~(\Cref{app:designer:prompts:validation-refine}).

\paragraph{Compact Summary} \myalgo{} maintains a design log to track a compact summary of the design process. The design log is updated every generation after obtaining feedback metrics on the new designs. The system message for \textbf{Compact Summary} is provided in \Cref{app:designer:prompts:system-compact}. The LLM is prompted to maintain a concise summary of the strategies explored, Pareto trade-offs observed, and open directions to explore, to guide the overall design process.

\inputcbminted[colframe=blue!50!black,coltitle=white,fontupper=\ttfamily\footnotesize]{prompts/system.md}{markdown}{Refine Design (System Prompt)}{app:designer:prompts:system-refine}
\inputcbminted[colframe=red!60!black,coltitle=white,fontupper=\ttfamily\footnotesize]{prompts/obj_improve.md}{markdown}{Improve Objective (Refine Design)}{app:designer:prompts:obj-improve}
\inputcbminted[colframe=green!50!blue,coltitle=white,fontupper=\ttfamily\footnotesize]{prompts/obj_explore.md}{markdown}{Explore Objective (Refine Design)}{app:designer:prompts:obj-explore}

\noindent
\begin{minipage}[t]{0.495\textwidth}
    \inputcbminted[colframe=gray!20!white,coltitle=black,fontupper=\ttfamily\footnotesize]{prompts/user.md}{markdown}{User Msg (Refine Design)}{app:designer:prompts:user-refine}
\end{minipage}\hfill
\begin{minipage}[t]{0.495\textwidth}
    \inputcbminted[colframe=gray!20!white,coltitle=black,fontupper=\ttfamily\footnotesize]{prompts/insp.md}{markdown}{Inspiration (Refine Design)}{app:designer:prompts:insp-refine}
\end{minipage}
\inputcbminted[colframe=gray!20!white,coltitle=black,fontupper=\ttfamily\footnotesize]{prompts/validation.md}{markdown}{Validation Retry (Refine Design)}{app:designer:prompts:validation-refine}

\inputcbminted[colframe=red!50!blue,coltitle=white,fontupper=\ttfamily\footnotesize]{prompts/sys_compact.md}{markdown}{Compact Summary (System Prompt)}{app:designer:prompts:system-compact}
\inputcbminted[colframe=gray!20!white,coltitle=black,fontupper=\ttfamily\footnotesize]{prompts/template.py}{python}{Design Template (\task{point-to-disk})}{app:designer:prompts:template}

\subsection{\myalgo\ : Designer Loop}
\label{app:designer:loop}
\myalgo\ utilizes a generic genetic algorithm~(GA) loop that utilizes a feasible-first~\citep{deb2000efficient} tournament selection~\citep{goldberg1991comparative,miller1995genetic}.
The crossover operator is handled by the LLM by suitable \ssf{RefineDesign} prompts (outlined in \Cref{app:designer:prompts}).
The refinement process starts with a few monolithic seed designs that characterize the trade-offs involved in monolithic configurations. The population size for the GA loop is capped to $N$ designs: all $N_{\text{new}}$ offspring and top $N - N_{\text{new}}$ designs from the previous generation are retained. The overall design loop of \myalgo\ is described in \Cref{alg:myalgo}.
Across tasks, we use a common set of hyperparameters for the design loop and they are outlined in \Cref{tab:design-loop:hyperparameters}. For all our experiments we use the \texttt{gemini-3.1-pro} reasoning LLM from Google DeepMind for its strong instruction following and code generation capabilities. We use the generic ``\texttt{chat completion}'' method exposed in the \texttt{any-llm} library~\citep{any-llm} to interface with the Gemini API, and resort to default sampling parameters (e.g. temperature is set to $1.0$) -- we leave optimizing prompts, sampling, and model selection for future work.

\begin{algorithm}[H]
\caption{\small \myalgo: \myalgoexpanded}
\label{alg:myalgo}
\small
\textbf{Input:} MDP $\gM$, Sensor Library $\Phi$, Cost $C: \Phi \to \R^+$, Synthesis Procedure $\mathsf{M}$, Seed Population: $\mathsf{D}$ \\[0.3em]
\textbf{Parameters:} \\
$N$ (population size), $G$ (generations), $N_{\text{new}}$ (\#offspring), $T$ (tournament size), \\[0.2em]
$\eps$ (performance threshold), $K$ (\#inspirations), $\rho$ (explore rate) \vspace{0.5em}
\begin{algorithmic}[1]
    \setlength{\itemsep}{0.2em}
    \Procedure{Select}{$\mathsf{D}$}
        \Comment{\small Feasible-first Rule~\citep{deb2000efficient}}
        \State Sample $T$ designs from $\mathsf{D}$; rank by $\big(\max(\eps - \mathrm{J}(\gD), 0),\, \mathrm{R}(\gD)\big)$; \Return best.
    \EndProcedure
    \vspace{0.5em}

    \Procedure{RefineDesign}{$\gD_{\text{curr}}$, $\mD_{\text{insp}}$, $\mathrm{L}$, $\text{objective}$}
    \State prompt $\gets$ \textsf{sys\_msg}($\gM,\,\Phi,\,C,\,\text{objective}$)
    \Comment{\Cref{app:designer:prompts:system-refine,app:designer:prompts:obj-improve,app:designer:prompts:obj-explore}}
    \State prompt $\gets$ prompt + \textsf{user\_msg}($\gD_{\text{curr}},\,\mD_{\text{insp}},\,\mathrm{L}$)
    \Comment{\Cref{app:designer:prompts:user-refine,app:designer:prompts:insp-refine}}
    \State retry $\gets$ 0
    \While{retry $<$ \texttt{max\_retries}}
    \State $\gD_{\text{new}} \gets$ \textsf{LLM}(prompt)
        \If{isValid($\gD_{\text{new}}$)}
            \State \Return $\gD_{\text{new}}$
        \Else
            \State prompt $\gets$ prompt + \textsf{validation\_msg}(error)
            \State retry $\gets$ retry + 1
        \EndIf
    \EndWhile
    \EndProcedure
    \vspace{0.5em}
    \Procedure{CompactSummary}{$\mathrm{L}$, $\mathsf{D}_{\text{new}}$}
    \State prompt $\gets$ \textsf{compact\_sys\_msg}($\gM,\,\Phi,\,C$) + \textsf{compact\_user\_msg}($\mathrm{L},\,\mathsf{D}_{\text{new}}$)
    \Comment{\Cref{app:designer:prompts:system-compact}}
    \State $\mathrm{L}_{\text{new}} \gets$ \textsf{LLM}(prompt)
    \State \Return $\mathrm{L}_{\text{new}}$
    \EndProcedure
    \vspace{1em} {\color[RGB]{80, 80, 80}\hrule} \vspace{1em}
    \Statex {\color[RGB]{80, 170, 80}\texttt{// \myalgo\ Main Loop}}
    \State $\mathsf{D}_0 \gets \mathsf{D}$
    \State $\mathrm{L} \gets \Call{CompactSummary}{\emptyset,\,\mathsf{D}_0}$
    \For{$g = 1$ \textbf{to} $G$}
        \State $\mathsf{D}_{\textsf{new}} \gets \emptyset$
        \Statex \hspace{\algorithmicindent} {\color[RGB]{80, 170, 80}\texttt{// Design Generation}}
        \For{$n = 1$ \textbf{to} $N_{\text{new}}$} %
            \State $\mD \gets $ run $\Call{Select}{\mathsf{D}_{g-1}}$ until $K + 1$ distinct designs are selected.
            \State $\gD_\text{curr}, *\mD_\text{insp} \gets \mD$
            \State $\text{objective} \gets \text{``explore''}$ if $\text{rand}() < \rho$ else $\text{``improve''}$
            \State $\gD_{\textsf{new}} \gets \Call{RefineDesign}{\gD_\text{curr}, \mD_\text{insp}, \mathrm{L}, \text{objective}}$
            \State $\mathsf{D}_{\text{new}} \gets \mathsf{D}_{\text{new}} \cup \{\gD_{\text{new}}\}$
        \EndFor
        \vspace{0.3em}
        \Statex \hspace{\algorithmicindent} {\color[RGB]{80, 170, 80}\texttt{// Synthesis + Evaluation}}
        \For{$\gD_i \in \mathsf{D}_{\textsf{new}}$} %
            \State $\Pi_{\gD_i} \gets \mathsf{M}(\gD_i)$;\enspace estimate $\mathrm{R}(\gD_i)$ and $\mathrm{J}(\gD_i)$
        \EndFor
        \vspace{0.3em}
        \Statex \hspace{\algorithmicindent} {\color[RGB]{80, 170, 80}\texttt{// Update Design Log}}
        \State $\mathrm{L} \gets \Call{CompactSummary}{\mathrm{L},\,\mathsf{D}_{\textsf{new}}}$ %
        \vspace{0.5em}
        \Statex \hspace{\algorithmicindent} {\color[RGB]{80, 170, 80}\texttt{// Update Active Population and Design Archive}}
        \State $\mD \gets$ select top $(N - N_{\text{new}})$ designs from $\mathsf{D}_{g-1}$ ordered by $\big(\max(\eps - \mathrm{J}(\gD), 0),\, \mathrm{R}(\gD)\big)$
        \State $\mathsf{D}_g \gets \mD \cup \mathsf{D}_{\textsf{new}}$;\enspace $\mathsf{D} \gets \mathsf{D} \cup
        \mathsf{D}_{\textsf{new}}$ %
    \EndFor \vspace{0.5em}
    \State \Return $\gD_\eps^* = \argmin_{\gD \in \mathsf{D}} \big(\max(\eps - \mathrm{J}(\gD), 0),\, \mathrm{R}(\gD)\big)$
\end{algorithmic}
\end{algorithm}

\begin{table}
    \centering
    {
    \renewcommand{\arraystretch}{1.2}
    \begin{tabular}{llc}
        \multicolumn{2}{l}{\bf Hyperparameters} & {\bf Value} \\ \hline
        $N$ & Population Size & 20 \\
        $N_{\text{new}}$ & Offspring per Generation & 3 \\
        $T$ & Tournament Size & 5 \\
        $K$ & Number of Inspirations & 3 \\
        $\rho$ & Explore Rate & 0.2 \\
               & \texttt{max\_retries} (LLM Validation) & 1 \\[1em]
    \end{tabular}
    }
    \caption{The hyperparameters of \myalgo{} designer loop across tasks.}
    \label{tab:design-loop:hyperparameters}
\end{table}

\section{\task{point-to-disk}}
\label{app:p2d}
\paragraph{Task} A velocity-controlled point-mass with state $\rvx$ (cartesian position) must be driven into a target disk of radius $r^* = 0.01$ at the origin and held there.
Given a velocity command $\rvu$, the system evolves as $\dot{\rvx} = \rvv$ with $\rvv \sim B_\epsilon(\rvu) := \{ \vv : ||\vv - \rvu||_2 \leq \eps ||\rvu||_2 \}$ and $\eps = 0.25$ -- the actuation noise grows with the commanded speed, so slowing down ($\rvu \to \mathbf{0}$) allows the controller to suppress actuation noise.
The controller, operating at 50\,Hz, is scored by its dwell time inside the target for rollouts of length $H = 500$ steps (10\,s) from states on the unit circle,
$J(\pi) = \E_{\pi,\,\{\rvx_0 : ||\rvx_0||_2 = 1\}}\left[ \sum_{t=0}^H \Ind{||\rvx_t||_2 < r^*} \right]$.
The designer's goal is to meet a performance target $\mathrm{J}_\text{target}$ while minimizing the sensing cost incurred en route; the sensor library and cost model used in the experiments~(\Cref{sec:results}) are detailed below.

\begin{wrapfigure}{r}{0.5\textwidth}
    \centering
    \vspace{-\baselineskip}
    \begin{minipage}{0.32\linewidth}
      \centering
      \includegraphics[width=\linewidth]{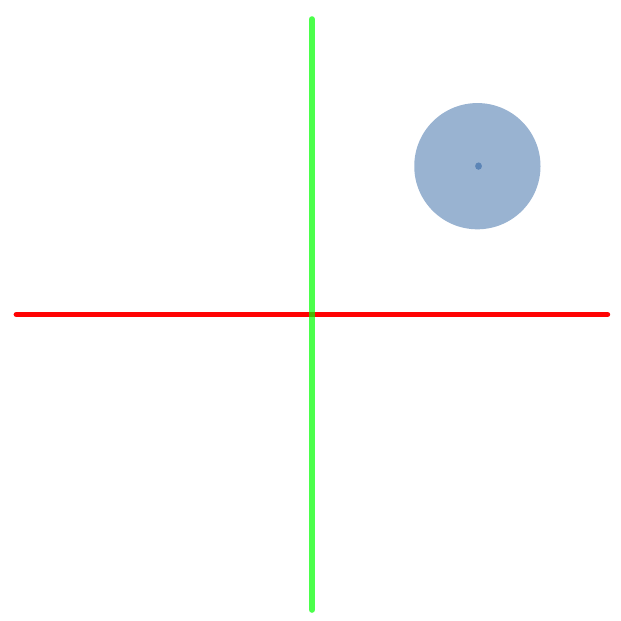}\\
      \stt{Cartesian}
    \end{minipage}
    \begin{minipage}{0.32\linewidth}
      \centering
      \includegraphics[width=\linewidth]{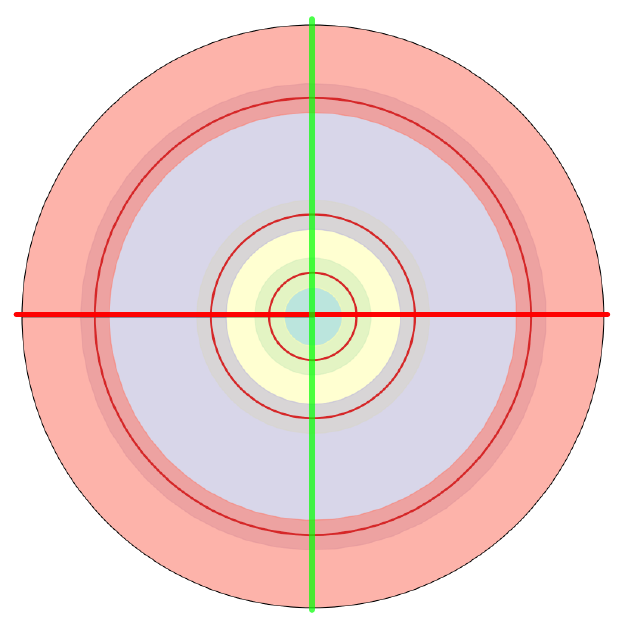}\\
      \stt{Radial}
    \end{minipage}
    \begin{minipage}{0.32\linewidth}
      \centering
      \includegraphics[width=\linewidth]{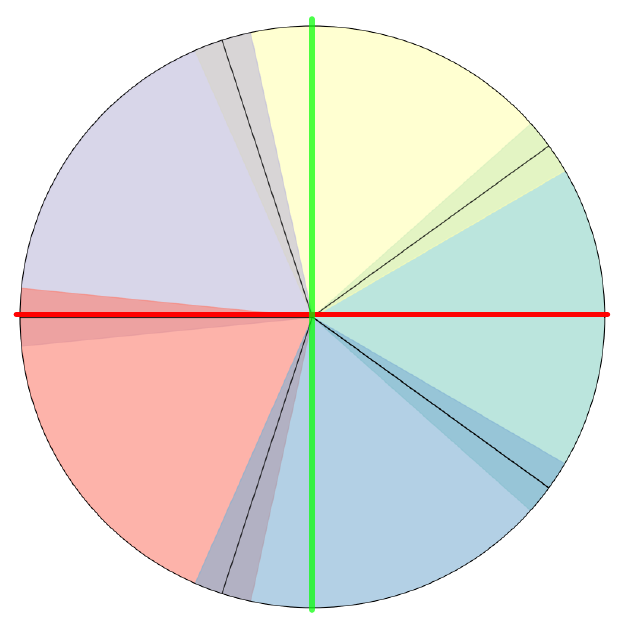}\\
      \stt{Sector}
    \end{minipage}
    \vspace{-\baselineskip}
\end{wrapfigure}
\paragraph{Sensors}
The designer can equip each mode with one or more of three sensor types, each revealing a different slice of the robot's position $\rvx$. In each mode, the agent's observation is the concatenation of the active sensors' readings (optionally stacked over a history length $h$), and pays the sum of their per-step costs.
\begin{itemize}[leftmargin=1.2em,itemsep=0.2em]
  \item \stt{Cartesian} (\texttt{GPS}): noisy position $\rvx + \gN(0, \sigma_\text{xy}^2 \mI_2)$, with $\sigma_\text{xy} \in \{0,\,10^{-2},\,5{\cdot}10^{-2},\,10^{-1},\,5{\cdot}10^{-1}\}$.
  \item \stt{Radial} \& \stt{Sector} (\stt{RadialSector}): a discretized polar reading returning the active sector among $s \in [1,360]$ equiangular sectors (optionally offset by $\phi$ radians) and the active radial band among $\text{len}(\rvd){+}1$ bands defined by increasing distance thresholds $\rvd = (d_1, \dots) \le 1$.
    The per-step output is a concatenation of the one-hot over radial bands (empty when $\rvd = ()$) and unit-vector $(\cos\theta_s,\sin\theta_s)$ corresponding to centroid of the active sector.
    There is some Gaussian noise in measuring the polar coordinates of the robot $(r, \theta)$. The noise scales are tunable with $\sigma_\theta \in \{0,\,10^{-1},\,2{\cdot}10^{-1},\,4{\cdot}10^{-1},\,8{\cdot}10^{-1}\}$ and $\sigma_r \in \{0,\,10^{-2},\,5{\cdot}10^{-2},\,10^{-1}\}$.
\end{itemize}

\subsection{Cost Model}
\label{app:p2d:cost}
We adopt an information-theoretic per-step cost that prices a sensor configuration $\varphi$ by the number of bits it resolves about the state.
For a quantity confined to a range ``$\text{span}$'' and recovered with effective standard deviation ``$\sigma$'', we use the resolution heuristic: $\text{bits}(\text{span},\sigma) \approx \max\!\big(\log_2(\text{span}/\sigma),\,0\big)$, i.e. the bits needed to index the $\approx\!\text{span}/\sigma$ distinguishable $\sigma$-wide cells tiling the range.
The per-step cost includes a fixed activation term, a term that scales with the bits resolved along each measured axis, and a storage term as follows:
\begin{align*}
  C_\text{radar}(\varphi) &= c^\text{rad}_\text{base} + c_\text{ang}\cdot\text{bits}(2\pi, \tilde\sigma_\theta) + c_\text{dist}\cdot\text{bits}(r_{\max}, \tilde\sigma_r) + \underbrace{c_\text{mem}\log_2 h}_{\text{storage}}, \\
  C_\text{gps}(\varphi) &= c^\text{gps}_\text{base} + 2\,c_\text{gps}\cdot\text{bits}(2 r_{\max}, \sigma_\text{xy}) + c_\text{mem}\log_2 h,
\end{align*}
where the GPS factor of $2 c_\text{gps}$ accounts for the two cartesian axes (each spanning $[-r_{\max}, r_{\max}]$).

\paragraph{Effective noise from quantization}
A \stt{RadialSector} resolves the state through both its measurement noise \emph{and} its discretization, so we convert each discretization into an equivalent variance and add it to the measurement variance before estimating the bits: $\tilde\sigma = \sqrt{\sigma^2_\text{quant} + \sigma^2_\text{meas}}$.
For the \emph{angular} channel, knowing only which of the $s$ equiangular sectors contains the heading vector results in a uniform spread of possible locations over an arc of width $w = 2\pi/s$.
This spread is measured by the mean chord length $R = \sin(w/2)/(w/2)$, with $R{=}1$ a perfectly localized heading and $R{=}0$ presenting no information about the heading.
This results in an equivalent quantization variance $\sigma^2_{\text{quant},\theta} = -2\ln R$, which is added to the measurement variance $\sigma_\theta^2$ to get the effective noise $\tilde\sigma_\theta$ for the angular channel.
For the \emph{radial} channel, bands with edges $0 = d_0 < d_1 < \dots < d_n < d_{n+1} = r_{\max}$ localize the radial distance only to the band it falls in; a uniform spread across a band of width $w_i = d_i - d_{i-1}$ contributes variance $w_i^2/12$. Weighting each band by the mass an annulus receives under a spatially-uniform prior on the disk, $p_i = (d_i^2 - d_{i-1}^2)/r_{\max}^2$, gives $\sigma^2_{\text{quant},r} = \sum_i p_i\, w_i^2/12$ and $\tilde\sigma_r = \sqrt{\sigma^2_{\text{quant},r} + \sigma_r^2}$.
With $\rvd = ()$, the radial channel is uninformative and its cost is dropped. Similarly, with $s=1$, the angular channel becomes uninformative and is dropped from the cost.
This cost specification exposes a smooth, monotone cost function over the parameter space of \stt{Sensor} configurations, which the designer can navigate to trade off cost and performance.

\paragraph{History as storage cost, not re-acquisition}
A mode that stacks $h$ readings already pays the full per-step acquisition cost (the bits above) at \emph{every} step, so buffering past observations reveals no information the sensor stream has not already been charged for. Its only marginal cost is the memory to retain $h$ readings, which we price as the sub-linear $c_\text{mem}\log_2 h$ -- reflecting the addressing/storage overhead of a depth-$h$ buffer rather than charging $h$-fold the acquisition cost.

\paragraph{Cost Structures}
We instantiate this model with two coefficient sets that encode contrasting hardware regimes -- \ssf{CostlyGPS} and \ssf{CheapGPS}. Both share $r_{\max}=2$ and $c_\text{mem}=0.005$.
\begin{table}[H]
  \centering\small
  {\renewcommand{\arraystretch}{1.2}
  \begin{tabular}{lccccc}
    \multirow{2}{*}{\bf Cost Structure} & \multicolumn{2}{c}{\sf GPS} & \multicolumn{3}{c}{\sf RadialSector} \\
                                        & $c^\text{gps}_\text{base}$ & $c_\text{gps}$ & $c^\text{rad}_\text{base}$ & $c_\text{ang}$ & $c_\text{dist}$ \\[0.2em] \hline
    \vspace{0.2em}\ssf{CostlyGPS} & $0.01$ & $0.01$ & $0.01$ & $0.01$ & $0.01$ \\
    \ssf{CheapGPS}  & $0.02$ & $0.01$ & $0.01$ & $0.05$ & $0.01$ \\
  \end{tabular}}
\end{table}
\noindent Under \ssf{CostlyGPS}, the radial and angular bits are cheap, so \stt{RadialSector} sensors dominate the affordable end of the spectrum. In \ssf{CheapGPS}, the angular coefficient is increased by $5\times$, making well-chosen \stt{Cartesian} configurations cost-competitive with \stt{RadialSector}. \Cref{tab:p2d:costs} reports representative per-step costs under each structure.
\begin{table}[H]
  \centering\small
  {\renewcommand{\arraystretch}{1.2}
  \begin{tabular}{lcc}
    {\bf Sensor configuration} & \ssf{CostlyGPS} & \ssf{CheapGPS} \\ \hline
    \texttt{SectorRadar(s=1, d=())} & $0.010$ & $0.010$ \\
    \texttt{SectorRadar(s=3, d=())} & $0.044$ & $0.178$ \\
    \texttt{SectorRadar(s=4, d=(0.01,0.05))} & $0.066$ & $0.217$ \\
    \texttt{SectorRadar(s=8, d=(0.05,0.2,0.5))} & $0.080$ & $0.272$ \\
    \texttt{GPS(sigma\_xy=0.01)} & $0.183$ & $0.193$ \\
    \texttt{GPS(sigma\_xy=0.05)} & $0.136$ & $0.146$ \\
    \texttt{GPS(sigma\_xy=0.1)} & $0.116$ & $0.126$ \\[1em]
  \end{tabular}}
  \caption{Representative per-step sensor costs under the two cost structures.}
  \label{tab:p2d:costs}
\end{table}

\subsection{Experimental Setup}
\label{app:p2d:setup}
A design $\gD$ is a \texttt{Python} program specifying \texttt{mode\_obs} (the sensor configuration per mode), \texttt{mode\_transitions} (predicates over the position $\rvx$), and \texttt{initial\_mode} (see the template in \Cref{app:designer:prompts:template}). Each candidate is synthesized by running \texttt{MSC-PPO} (\Cref{app:msc-rl}, hyperparameters in \Cref{tab:msc-ppo:hyperparameters}) with 3 random seeds; the privileged critic and monitors operate on the polar state $(\|\rvx\|, \cos\theta, \sin\theta)$. The oracle controller with perfect state achieves $\mathrm{J}_\text{oracle} = 456.23 \pm 4.59$ ($N=1000$). We set performance targets for \myalgo\ to $\mathrm{J}_\text{target} \in \{200, 300, 450\}$ ($\approx 0.44, 0.66, 0.98$ of oracle performance), spanning aggressive to near-lossless performance budgets.

We seed the search with $10$ monolithic configurations~(see \Cref{tab:p2d:seeds}) -- containing $7$ \stt{RadialSector} configurations spanning sector counts and angular-noise levels and $3$ \stt{Cartesian} configurations -- and run \myalgo\ for $8$ generations under each (cost structure, performance target) pair.
These seed designs trace out an approximate monolithic frontier, and the same seed population is used for both the cost structures.
\myalgo\ is provided with the code implementing the task and cost function as part of the system prompt.

\begin{table}[H]
  \centering\small
  {\renewcommand{\arraystretch}{1.2}
  \begin{tabular}{lccc}
    & & \multicolumn{2}{c}{\bf Per-step cost} \\
      {\bf Configuration} & {\bf $\mathrm{J}/\mathrm{J}_\text{oracle}$} & \ssf{CostlyGPS} & \ssf{CheapGPS} \\[0.2em] \hline
    \multicolumn{4}{l}{\sffamily\footnotesize\underline{\texttt{RadialSector} monoliths}} \\
    \texttt{SectorRadar(s=3, sigma\_theta=0.8)}     & $0.39\,\pm\,0.03$ & $0.036$ & $0.142$ \\
    \texttt{SectorRadar(s=3)}                       & $0.43\,\pm\,0.03$ & $0.044$ & $0.178$ \\
    \texttt{SectorRadar(s=4)}                       & $0.66\,\pm\,0.03$ & $0.048$ & $0.199$ \\
    \texttt{SectorRadar(s=5, sigma\_theta=0.2)}     & $0.76\,\pm\,0.02$ & $0.049$ & $0.206$ \\
    \texttt{SectorRadar(s=7)}                       & $0.80\,\pm\,0.02$ & $0.056$ & $0.240$ \\
    \texttt{SectorRadar(s=10, sigma\_theta=0.2)}    & $0.83\,\pm\,0.02$ & $0.055$ & $0.237$ \\
    \texttt{SectorRadar(s=360)}                     & $0.87\,\pm\,0.01$ & $0.113$ & $0.524$ \\[0.2em]
    \multicolumn{4}{l}{\sffamily\footnotesize\underline{\texttt{Cartesian} monoliths}} \\
    \texttt{GPS(sigma\_xy=0.05, history=5)}         & $0.49\,\pm\,0.08$ & $0.148$ & $0.158$ \\
    \texttt{GPS(sigma\_xy=0.01)}                    & $0.98\,\pm\,0.01$ & $0.183$ & $0.193$ \\
    \texttt{GPS(sigma\_xy=$10^{-6}$)}               & $1.00\,\pm\,0.01$ & $0.449$ & $0.459$ \\[1em]
  \end{tabular}}
  \caption{The monolithic seed designs used to initialize \myalgo\ for \task{point-to-disk}.}
  \label{tab:p2d:seeds}
\end{table}

\section{\task{RF-DMC} (RangeFinder Variants of \texttt{dm\_control})}
\label{app:dmc}
We construct rangefinder variants of three \texttt{dm\_control}~\citep{tunyasuvunakool2020,mujoco_playground_2025} tasks -- \task{cartpole-swingup}, \task{cup-catch}, and \task{finger-spin}. In each, the agent observes the world through a RayScanSensor consisting of $360^\circ$ rangefinders that are rigidly attached to a task-relevant body site (pole tip, cup center, fingertip) and cast equiangular rays that report the distance to the nearest enclosing-box surface (we wrap each scene in walls/ceiling so every ray returns a finite range).
For each mode, the designer configures a \texttt{RayScanSensor} by choosing its measurement quality $q\in\{\mathsf{A},\dots,\mathsf{F}\}$, the number of equispaced rays $n$ (a divisor of $360$), an optional angular offset $\phi$, and a history length $h$.
The mode-transition predicates $\delta$ and initial-mode map $\nu$ are privileged: they read interpretable simulator state (e.g. \texttt{pole\_angle}, \texttt{ball\_to\_target}, \texttt{hinge\_angvel}) and these transitions are distilled into rangefinder-based monitors by \texttt{MSC-PPO} (\Cref{app:msc-rl}).
Task performance is the oracle-normalized episodic return, $\mathrm{J}/\mathrm{J}_\text{oracle}$, estimated over $N{=}1000$ rollouts.

\paragraph{Tasks} All three tasks run for $H=1000$ steps with a per-step reward in $[0,1]$.
\begin{itemize}[leftmargin=1.2em,itemsep=0.2em]
    \item \task{cartpole-swingup} (100\,Hz): a force-controlled cart must swing up a hinged pole and balance it upright near the center of the rail. The reward multiplicatively combines uprightness, centering, small control, and small angular velocity. To ensure the task is solvable by \texttt{MSC-PPO} we use a wider reset distribution during training by spawning the pole roughly upright 50\% of the time, but preserve the standard reset distribution during evaluation.
    \item \task{cup-catch} (50\,Hz): a force-actuated cup must catch a ball tethered to it by a $0.3$\,m rope, possibly swinging it up first. The reward is the indicator function checking if the ball is successfully caught.
    \item \task{finger-spin} (50\,Hz): a two-link finger must repeatedly strike a free-hinged spinner to maintain an angular velocity ($>15$\,rad/s). The reward is the indicator function checking if the spinner's angular velocity exceeds the target. For \texttt{MSC-PPO} training we use a smoother reward function that provides a dense signal to get the spinner spinning at a rate higher than the target. However, the original binary reward is preserved during controller evaluation.
\end{itemize}
\Cref{tab:dmc:tasks} summarizes the per-task settings, oracle performance, and the performance target used for \myalgo\ in the experiments presented in \Cref{sec:results}.

\begin{table}[H]
  \centering\small
  {\renewcommand{\arraystretch}{1.2}
  \begin{tabular}{llccc}
      {\bf Task} & {\bf Rangefinder Site} & {\bf Priv.\ State Size} & {$\mathrm{J}_\text{oracle}$} & {$\mathrm{J}_\text{target}\,(/\mathrm{J}_\text{oracle})$} \\ \hline
    \task{cartpole-swingup} & pole tip   & $5$  & $881.8 \pm\:\, 4.9$  & $650\;(0.74)$ \\
    \task{cup-catch}        & cup center & $10$ & $977.8 \pm 15.7$ & $700\;(0.72)$ \\
    \task{finger-spin}      & fingertip  & $19$ & $969.9 \pm 11.7$ & $750\;(0.77)$ \\[1em]
  \end{tabular}}
  \caption{\task{RF-DMC} task settings. The oracle performance corresponds to episodic return achieved with full state information (measured over $N = 1000$ rollouts).}
  \label{tab:dmc:tasks}
\end{table}

\subsection{Rangefinder Sensor Model}
\label{app:dmc:sensor}
We adopt the probabilistic beam-sensor model of \citet{thrun2005probabilistic}, a well-characterized non-Gaussian noise model that abstracts noisy exteroceptive sensors (e.g. LiDAR) while remaining fast to simulate. A measured range $z$ given the true range $z^*$ is drawn from a four-component mixture capturing: measurement noise, unexpected short returns, detection failures, and spurious returns.
\[
  p(z \mid z^*) = c_{\text{hit}}\,p_{\text{hit}}(z) + c_{\text{short}}\,p_{\text{short}}(z) + c_{\text{max}}\,p_{\text{max}}(z) + c_{\text{rand}}\,p_{\text{rand}}(z),\quad \textstyle\sum_* c_* = 1,
\]
Where:
\begin{itemize}[leftmargin=1.2em,itemsep=0.2em]
    \item $p_{\text{hit}} = \mathcal{N}_{[z_{\min},z_{\max}]}(z^*,\sigma_{\text{hit}}^2)$ (truncated Gaussian), models the inherent measurement noise.
    \item $p_{\text{short}} = \mathrm{Exp}_{[z_{\min},z^*]}(\lambda_{\text{short}})$ (truncated exponential), models unexpected interference.
  \item $p_{\text{max}} = \delta_{z=z_{\max}}$, models detection failures.
  \item $p_{\text{rand}} = \mathrm{Uniform}(z_{\min},z_{\max})$, models spurious readings.
\end{itemize}
Each measurement is quantized to resolution $z_{\text{res}}$ and normalized to $[0,1]$ by $z_{\max}$.
The discrete quality grades $\mathsf{A}$ (cleanest) through $\mathsf{F}$ (noisiest) instantiate distinct mixture weights and $\sigma_{\text{hit}}$ (all share $c_{\text{short}} = 0$, $z_{\text{res}} = 10^{-5}$, $z_{\max} = 10$), and are summarized in \Cref{tab:dmc:grades}.

Each grade's per-measurement energy is derived from its $90$th-percentile signal-to-noise ratio $\mathrm{SNR}_\text{dB}$ over the sensor's operating range $[z_{\min}, z_{\max}]$ (estimated by Monte-Carlo sampling of the beam model) as:
\[
  \mathrm{E}(q) = k_s\, z_{\max}^{4}\,\mathrm{SNR}_\text{dB}^3 \;+\; k_c\, z_{\text{res}}^{-1}, \qquad k_s = 10^{-2},\; k_c = 10^{-3},
\]
where the first term models acquisition energy (higher quality requires substantially more energy) and the second a fixed representation/compute overhead.
\Cref{tab:dmc:grades} reports these energies normalized to grade $\mathsf{F}$, which serve as the per-ray costs in the cost model described below.

\begin{table}[H]
  \centering\small
  {\renewcommand{\arraystretch}{1.2}
  \begin{tabular}{crrrrrr}
    {\bf Grade} & $c_{\text{hit}}$ & $c_{\text{max}}$ & $c_{\text{rand}}$ & $\sigma_{\text{hit}}$ & $\mathrm{SNR}_\text{dB}$ & {\bf Rel.\ Cost / Ray} \\ \hline
    $\mathsf{A}$ & $0.995$ & $0.003$ & $0.002$ & $10^{-5}$ & $32.0$ & $22.61$ \\
    $\mathsf{B}$ & $0.990$ & $0.005$ & $0.005$ & $8{\cdot}10^{-4}$ & $28.1$ & $15.39$ \\
    $\mathsf{C}$ & $0.970$ & $0.010$ & $0.020$ & $10^{-3}$ & $22.2$ & $7.62$ \\
    $\mathsf{D}$ & $0.950$ & $0.010$ & $0.040$ & $10^{-2}$ & $19.4$ & $5.08$ \\
    $\mathsf{E}$ & $0.850$ & $0.050$ & $0.100$ & $5{\cdot}10^{-2}$ & $15.2$ & $2.44$ \\
    $\mathsf{F}$ & $0.650$ & $0.100$ & $0.250$ & $10^{-1}$ & $11.3$ & $1.00$ \\[1em]
  \end{tabular}}
  \caption{Rangefinder quality grades: mixture weights, hit-noise scale, $90$th-percentile SNR, and per-ray energy cost normalized to grade $\mathsf{F}$ (all grades use $c_{\text{short}}{=}0$).}
  \label{tab:dmc:grades}
\end{table}

\subsection{Cost Model}
\label{app:dmc:cost}
The per-step cost of a \texttt{RayScanSensor} scales the per-ray energy by the number of rays and adds a logarithmic memory term for the history buffer:
\[
  C(\varphi) = \big(\mathrm{E}(q) + c_\text{mem}\log_2 h\big)\cdot n, \qquad c_\text{mem} = 0.5,
\]
with $\mathrm{E}(q)$ the normalized per-ray cost from \Cref{tab:dmc:grades}. This couples the three knobs the way physical scanning rangefinders behave: cost grows linearly in resolution $n$, increases with quality through $\mathrm{E}(q)$, and pays only a sub-linear ($\log_2 h$) storage premium for buffering history (which re-uses already-acquired scans, cf.\ \Cref{app:p2d:cost}). A mode with an empty sensor list has zero cost. This cost model is shared across all the three tasks.

\subsection{Experimental Setup}
\label{app:dmc:setup}
A design $\gD$ specifies \texttt{mode\_obs} (a \texttt{RayScanSensor} configuration per mode), \texttt{mode\_transitions} (predicates over the privileged state), and \texttt{initial\_mode}; each candidate is synthesized with \texttt{MSC-PPO} (\Cref{app:msc-rl}, \Cref{tab:msc-ppo:hyperparameters}) using two seeds per design to lower training costs.
We seed \myalgo\ with $11$ monoliths (\Cref{tab:dmc:seeds}) and run it for $10$ generations. The seed designs are obtained by selecting RayScan configurations $(q, h, n)$ to uniformly span the cost spectrum.
The seed designs provide an approximate per-task monolithic frontier -- demonstrating the influence of resolution $n$, quality $q$, and history $h$ on performance -- and the same seed population is used for all three tasks.
\myalgo\ is provided with the code implementing the task and cost function as part of the system prompt.

\begin{table}[H]
  \centering\small
  {\renewcommand{\arraystretch}{1.2}
  \begin{tabular}{lrccc}
    & & \multicolumn{3}{c}{Task Performance:\enspace $\mathrm{J}/\mathrm{J}_\text{oracle}$} \\
    \cmidrule(lr){3-5}
    {\bf Config $(q,h,n)$} & {\bf cost/step} & \task{cartpole-swingup} & \task{cup-catch} & \task{finger-spin} \\ \hline
    $(\mathsf{F}, 4, 3)$     & $6.0$    & $0.26$ & $0.12$ & $0.00$ \\
    $(\mathsf{E}, 8, 3)$     & $11.8$   & $0.60$ & $0.11$ & $0.09$ \\
    $(\mathsf{D}, 16, 3)$    & $21.2$   & $0.96$ & $0.11$ & $0.37$ \\
    $(\mathsf{C}, 4, 4)$     & $34.5$   & $0.84$ & $0.82$ & $0.65$ \\
    $(\mathsf{C}, 4, 10)$    & $86.2$   & $0.93$ & $0.83$ & $0.94$ \\
    $(\mathsf{E}, 8, 60)$    & $236.6$  & $0.54$ & $0.22$ & $0.61$ \\
    $(\mathsf{C}, 8, 60)$    & $546.9$  & $0.94$ & $0.34$ & $0.95$ \\
    $(\mathsf{A}, 4, 60)$    & $1416.5$ & $0.96$ & $0.99$ & $0.99$ \\
    $(\mathsf{B}, 4, 180)$   & $2949.7$ & $0.92$ & $0.98$ & $0.99$ \\
    $(\mathsf{A}, 2, 360)$   & $8319.2$ & $0.80$ & $0.99$ & $0.99$ \\
    $(\mathsf{A}, 16, 360)$  & $8859.2$ & $0.98$ & $0.98$ & $0.87$ \\[1em]
  \end{tabular}}
  \caption{The $11$ monolithic seed designs used to initialize \myalgo\ across all \task{RF-DMC} tasks. Each seed design uses the same sensor \texttt{RayScanSensor(quality$=q$, history$=h$, nrays$=n$)} throughout the task. The task performance is measured over $N{=}1000$ rollouts.}
  \label{tab:dmc:seeds}
\end{table}

\section{\task{clutter-nav} (Cluttered Navigation with Radar Sensors)}
\label{app:nav}
A disk-shaped robot with unicycle (differential-drive) dynamics must reach a goal while avoiding static and dynamic disk obstacles in a cluttered planar world. Unlike \task{point-to-disk} and \task{RF-DMC}, the MSC controllers in this task are \emph{not} synthesized by \texttt{MSC-PPO}: instead each mode configures the Radar sensor and Model-Predictive Path-Integral (MPPI)~\citep{williams2015model} planner.
The MPPI planner acts under a naive certainty-equivalence assumption -- treating the noisy obstacle detections of the Radar as the ground truth state.
Additionally, the designer synthesizes mode-transition predicates \emph{directly} over the same noisy detections the planner sees -- thus, no controller learning or monitor distillation is required for this task.
This setup exercises \myalgo\ with a different synthesis procedure, performance metric, and a joint sensor-and-planner design space.

\paragraph{Task}
The robot is spawned in a $[-15, 15]^2$\,m plane. At reset, a random number of obstacles are spawned at uniformly random positions, drawing $n_\text{static}\!\in\![5,20]$ static and $n_\text{dynamic}\!\in\![0,15]$ dynamic obstacles with radii in $[0.3,2.0]$\,m.
The agent (radius $0.5$\,m) starts collision-free with a goal sampled $10$--$15$\,m away; an episode succeeds if the agent reaches within $\text{goal\_tol}=0.25$\,m of the goal before a $25$\,s ($H=500$ steps at $20$\,Hz, $dt=0.05$\,s) time limit.
At each timestep, the agent commands $(v,\omega)$ with $v\in[0,5]$\,m/s (forward velocity) and $\omega\in[-3,3]$\,rad/s (turning rate). There is actuation noise where the realized velocity is commanded velocity perturbed with a Gaussian noise that scales with the command magnitude ($\sigma=(0.1,0.1)$). The dynamic obstacles also operate with unicycle dynamics and follow a soft-repulsion-plus-goal-tracking policy towards periodically resampled goals.
When simulating this task the collision dynamics are disabled, so the agent and obstacles can interpenetrate. However, the number of such collisions is recorded and factored into the performance metric.
Task performance is measured by the pair $\mathrm{J} = (\text{success rate},\, 1/\mathrm{cpt}_{90})$, where the collision rate $\mathrm{cpt}_{90}$ is the $90$th-percentile over episodes of collisions $/$ steps.
\myalgo\ minimizes resource cost associated with sensor-planner configuration subject to a feasibility constraint of $\geq 80\%$ success while pushing $1/\mathrm{cpt}_{90}$ above a configured target threshold (\Cref{app:nav:setup}). We assume the agent always has access to a compass that reports the relative goal bearing and distance accurately.

\subsection{Radar Sensor Model}
\label{app:nav:sensor}
Each mode's Radar configuration returns ego-centric \texttt{Detections} -- range $r$, bearing $\mathrm{az}$, size $\mathrm{sz}$, and radial velocity $v_r$ -- for obstacles within \texttt{max\_range} and $|\mathrm{az}|\le \texttt{fov}/2$.
Detection reliability and noise are governed by an SNR proxy $P = (R_0/r)^4$: the detection probability is $p_\text{detect}=p_\text{fp}^{\,1/(1+P)}$ (near-certain for $r \ll R_0$, decaying to the false-positive rate beyond $R_0$),
and noise associated with different dimensions (range, bearing, size, velocity) scale as $\sigma_*(r) = \hat\sigma_*\,r^2/R_0$, with the base noise scales being $\hat\sigma_r = 0.2$\,m, $\hat\sigma_\text{az} = 0.05$\,rad, $\hat\sigma_\text{sz} = 0.08$\,m, $\hat\sigma_v = 0.3$\,m/s.
The detections are quantized into $1^\circ$ angular bins and return only the nearest obstacle per bin (respecting occlusions), $v_r$ captures only the radial component of obstacle motion.
Additionally, spurious detections (that appear as random blips in the ego-centric \texttt{Detections}) arrive at $\mathrm{Poisson}(1)$ per scan.
Thus larger $R_0$, \texttt{max\_range}, and \texttt{fov} enable cleaner, longer-range, and wider coverage but come at a higher cost.
Of the various knobs, the designer controls only $(R_0, \texttt{max\_range}, \texttt{fov})$; the noise scales and false-positive rate are held fixed across modes.

\subsection{MPPI Planner and Certainty Equivalence}
\label{app:nav:planner}
Each mode is driven by its own MPPI planner configuration acting under \emph{certainty equivalence}: it treats the latest noisy \texttt{Detections} as the true obstacle state and does not reason about sensing uncertainty.
The planner is warm-started with a shifted nominal plan, so on re-entering a mode (or right after a transition) it always refines the previous nominal plan rather than starting from scratch.
In a mode, MPPI re-plans once every \texttt{period} steps and commits to the plan for $\texttt{period}\,-\,1$ steps, amortizing both sensing and compute costs.

A plan is parameterized by $K=6$ control knots per action dimension $(v,\omega)$ in a normalized space $[-1,1]$. This is lifted to the $T=\texttt{horizon}/dt$-step control sequence by a fixed cubic Catmull–Rom spline basis and is affinely mapped to the actuation limits $v\in[0,5]$\,m/s and $\omega\in[-3,3]$\,rad/s, yielding smooth, low-dimensional plans.
Each re-plan first shifts the previous plan to seed an initial warmstarted plan, then runs \texttt{n\_iters} refinement iterations.
At iteration $i$ it samples \texttt{n\_samples} knot perturbations from a zero-mean Gaussian whose knot-space scale anneals geometrically from $\sigma_0 = 0.8$ to $\sigma_1 = 0.1$, $\sigma(i)= \sigma_0(\sigma_1/\sigma_0)^{i/(\texttt{n\_iters}\,-\,1)}$ (early iterations explore broadly, later iterations do local refinements).
The perturbed trajectories are rolled out and scored, to finally update the nominal knots to the soft-min weighted mean $w_\tau\propto\exp\!\big(-(\mathrm{cost}(\tau) -\min_{\tau'}\mathrm{cost}(\tau'))/\lambda\big)$ with temperature $\lambda = 10$.

Each rolled-out trajectory $\tau=\{p_t\}_{t=1}^{T}$ (ego-frame, starting at the origin) is scored against the ego-centric goal $g$ and the obstacles by a weighted sum of five terms,
\[
  \mathrm{cost}(\tau) = w_\text{hit}\,C_\text{hit} + w_\text{soft}\,C_\text{soft} + w_\text{goal}\,C_\text{goal} + w_\text{prog}\,C_\text{prog} + w_\text{ctrl}\,C_\text{ctrl},
\]
Let $d_{t,m}$ denote the clearance to obstacle $m$ at step $t$ (clearance measures the distance between the centers minus the agent+obstacle radii), then the five cost terms are:
\begin{itemize}[leftmargin=1.3em,itemsep=0.15em]
  \item $C_\text{goal}=\min_t\|p_t-g\|$ — closest approach to the goal over the horizon $(w_\text{goal}{=}5)$
  \item $C_\text{prog}=\operatorname{mean}_t \max(\|p_t-g\|-\rho,\,0)$ — mean residual distance rewarding steady progress $(w_\text{prog}{=}3)$
  \item $C_\text{hit}=\sum_t \Ind{\min_m d_{t,m}<0}$ — \# of horizon steps in collision $(w_\text{hit}{=}50)$
  \item $C_\text{soft}=\sum_t \exp\!\big(-(\max(\min_m d_{t,m},0)/\eta)^2\big)$ — a smooth proximity penalty with $\eta{=}0.1$\,m, disabled once the robot is $1\,$m from the goal $(w_\text{soft}{=}2)$
  \item $C_\text{ctrl}=\sum_k (v_k{+}1)^2+\omega_k^2$ — knot-space control effort, referenced so $v{=}0,\,\omega{=}0$ is the zero-cost minimum $(w_\text{ctrl}{=}1)$
\end{itemize}
When rolling out a plan, the obstacles are propagated forward with their observed radial velocity $v_r$ under a constant-velocity model, and the clearance $d_{t,m}$ is computed against these propagated obstacles.
Of the various knobs, the designer controls only $(\texttt{n\_samples},\texttt{n\_iters},\texttt{horizon},\texttt{period})$; everything else is held fixed across modes.

\begin{table}[H]
  \centering\small
  {\renewcommand{\arraystretch}{1.2}
  \begin{tabular}{cllc}
    & {\bf Parameter} & {\bf Description} & {\bf Domain} \\ \hline
    \multirow{3}{*}{\rotatebox{90}{\sf Sensor}}
    & $R_0$            & sensor power (reliable-detection range, $\propto$ SNR)     & $[2,\,50]$ \\
    & \texttt{max\_range} & range cutoff for returned detections                    & $[5,\,15]$ \\
    & \texttt{fov}     & angular field of view (rad); $2\pi$ omnidirectional        & $[\pi/3,\,2\pi]$ \\[0.1em]\hline
    \multirow{4}{*}{\rotatebox{90}{\sf Planner}}
    & \texttt{n\_samples} & MPPI rollouts per refinement iteration                  & $\{256,512,1024\}$ \\
    & \texttt{n\_iters} & MPPI refinement iterations                                & $\{1,\dots,5\}$ \\
    & \texttt{horizon} & planning horizon (s)                                       & $[2,\,5]$ \\
    & \texttt{period}  & re-plan \texttt{period}                                    & $\{1,\dots,20\}$ \\[1em]
  \end{tabular}}
  \caption{\task{clutter-nav} per-mode Sensor + Planner \texttt{Configuration} space.}
  \label{tab:nav:params}
\end{table}

\subsection{Design Space and Cost Model}
\label{app:nav:cost}
Each mode maps to a single \texttt{Configuration} jointly specifying Sensor and Planner parameters (\Cref{tab:nav:params}). The per-step cost sums a sensor (energy), coverage (area), and compute term, amortized by the re-plan period:
\[
    C(\varphi) = \frac{1}{p}\Big( w_s\underbrace{(R_0/R_{\text{ref}})^2}_{C_\text{sensor}} + w_c\underbrace{\tfrac{0.5\,\texttt{max\_range}^2\,\texttt{fov}}{A_\text{ref}}}_{C_\text{cov}} + w_p\underbrace{\tfrac{n\cdot i\cdot \texttt{horizon}}{F_\text{ref}}}_{C_\text{compute}} \Big),
\]
With sensor power measured relative to $R_\text{ref} = 2$, coverage to the wedge $A_\text{ref}=0.5\cdot 10^2\cdot(\pi/3)$, and the planner compute to the lowest rollout budget $F_\text{ref}=256\cdot 1\cdot 2.0$.
A larger re-plan period $p$ amortizes both sensing and compute at the risk of acting on stale plans.
We use weights $(w_s,w_c,w_p)=(1.0,\,0.5,\,1.0)$ for our experiments~(\Cref{sec:results}).
The resource cost spans three orders of magnitude across the design space, some examples are provided in \Cref{tab:nav:examples}.

\begin{table}[H]
  \centering\small
  {\renewcommand{\arraystretch}{1.2}
  \begin{tabular}{lr}
    {\bf Configuration} & {\bf cost/step} \\ \hline
    \texttt{(R0=2, max\_range=10, fov=$\pi$/3, n\_iters=1, n\_samples=256, horizon=2, period=10)}  & $0.25$ \\
    \texttt{(R0=20, max\_range=10, fov=$2\pi$, n\_iters=3, n\_samples=512, horizon=3, period=4)}   & $28.00$ \\
    \texttt{(R0=50, max\_range=15, fov=$2\pi$, n\_iters=5, n\_samples=1024, horizon=5, period=1)}  & $681.75$ \\[1em]
  \end{tabular}}
  \caption{Example per-step costs for \task{clutter-nav} under cost model described in \Cref{app:nav:cost}.}
  \label{tab:nav:examples}
\end{table}

\subsection{Experimental Setup}
\label{app:nav:setup}
The monolithic frontier is obtained by a dense sweep of $768$ \texttt{Configuration}s -- the grid $\texttt{n\_iters}\!\in\!\{1,3,5\}$, $R_0\!\in\!\{2,8,10,50\}$, $\texttt{max\_range}\!\in\!\{10,15\}$, $\texttt{n\_samples}\!\in\!\{256,1024\}$, $\texttt{fov}\!\in\!\{\pi/3,2\pi\}$, $\texttt{horizon}\!\in\!\{2,5\}$, $\texttt{period}\!\in\!\{1,4,8,10\}$. We seed \myalgo\ with $25$ of these monoliths spanning the entire cost range and run \myalgo\ up to $20$ generations for separate performance targets $1/\mathrm{cpt}_{90}\in\{30,40,50\}$ (i.e. $\mathrm{cpt}_{90}\le\{0.033,0.025,0.020\}$) under the $\geq 80\%$ success rate constraint. Every design is evaluated over $1000$ random starting configurations with a 500-step limit.

\section{Points of Comparison}
\label{app:compare}
\subsection{\texttt{BO-MSC}: BayesOpt on \task{point-to-disk}\ssf{[CostlyGPS]}}
\label{app:compare:bo}
When candidate evaluations are expensive, black-box optimizers such as Bayesian Optimization~(BayesOpt) are a natural tool-kit to consider. BayesOpt, however, can only tune the parameters of a \emph{fixed} design template; it cannot alter the program structure (the number of modes, sensor types, or transition logic) or otherwise synthesize qualitatively new strategies. To quantify this gap, we hand-author four MSC templates that hard-code structural motifs \myalgo\ synthesizes on \task{point-to-disk}\ssf{[CostlyGPS]}, and use BayesOpt to populate their free parameters (this combination is denoted \texttt{BO-MSC} in \Cref{fig:pareto}).

\begin{figure}[ht]
  \centering
  \includegraphics[width=\textwidth]{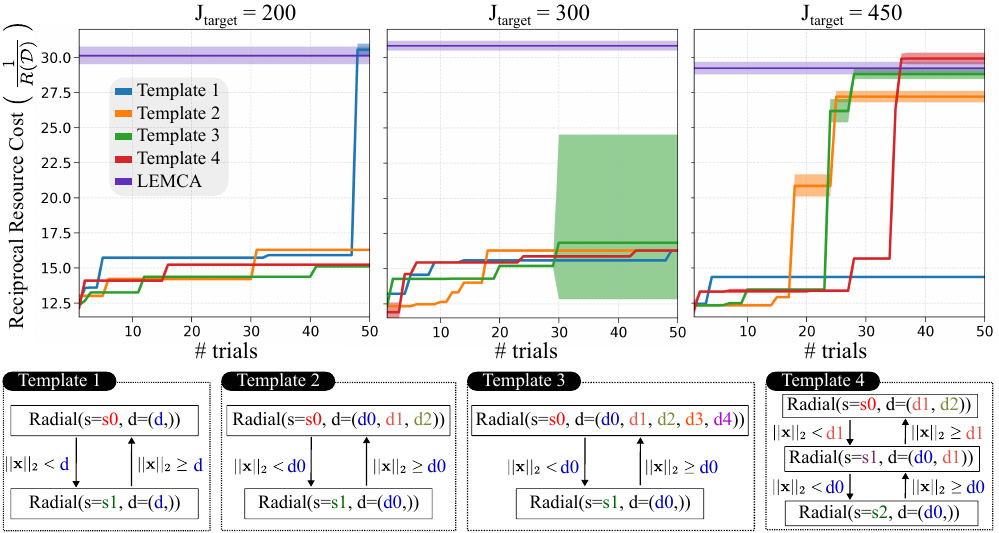}
  \caption{\texttt{BO-MSC} on \task{point-to-disk}\ssf{[CostlyGPS]}: we run black-box BayesOpt to populate the
  free parameters (highlighted in color) of four pre-specified mode-switching templates (\texttt{T1}--\texttt{T4}, bottom; see \Cref{tab:bo:templates}). Each panel plots the reciprocal resource cost of the best feasible parameterization found so far against the number of trials (evaluations), for a performance target $\mathrm{J}_\text{target}\!\in\!\{200,300,450\}$; the horizontal band is \myalgo's best design and shaded regions denote $\pm 1\sigma$ over 1000 rollouts.}
  \label{fig:p2d_result_BOpt}
\end{figure}

\paragraph{Templates.} Each template (\Cref{fig:p2d_result_BOpt}, bottom; \Cref{tab:bo:templates}) instantiates a nested ``target-disk'' strategy through \stt{RadialSector} detectors: the measured radius $\|\rvx\|_2$ partitions the state space into an inner mode covering the target disk and one or more outer ``approach'' modes. The templates progressively enrich the radial-band discretization available to the approach mode(s) -- supporting finer progressive braking as the robot nears the target -- while keeping the inner mode minimal. \texttt{T1} gives the approach mode a single radial band; \texttt{T2} and \texttt{T3} refine it to 3 and 5 bands; and \texttt{T4} instead splits the approach into two nested modes. For each template, BayesOpt tunes the per-mode angular \stt{Sector} resolution $s_i\!\in\![1,16]$ and the radial bands $d_j\!\in\![0,1]$, amounting to 3--7 integer/continuous parameters per template.

\begin{table}[H]
  \centering\small
  {\renewcommand{\arraystretch}{1.2}
  \begin{tabular}{cccl}
    \toprule
    {\bf Template} & {\bf Modes} & {\bf Free params} & {\bf Approach-mode radial structure} \\
    \midrule
    \texttt{T1} & 2 & 3 & single radial band \\
    \texttt{T2} & 2 & 5 & 3 radial bands \\
    \texttt{T3} & 2 & 7 & 5 radial bands \\
    \texttt{T4} & 3 & 6 & two nested approach modes \\[1em]
  \end{tabular}}
  \caption{The four hand-authored \texttt{BO-MSC} templates for \task{point-to-disk}\ssf{[CostlyGPS]}. All modes sense via \stt{RadialSector} detectors and transition based on $\|\rvx\|_2$; the free parameters are the per-mode \stt{Sector} resolutions $s_i$ and the radial band edges $d_j$ (\Cref{fig:p2d_result_BOpt}, bottom).}
  \label{tab:bo:templates}
\end{table}

\paragraph{Setup.} We use the Ax/BoTorch BayesOpt client~\citep{olson2025ax}, configured to minimize the resource cost $\mathrm{R}(\gD)$ subject to the outcome constraint $\mathrm{J}(\gD)\!\ge\!\mathrm{J}_\text{target}$ -- mirroring \myalgo's feasible-first objective. Crucially, every candidate is evaluated with the \emph{same} \texttt{MSC-PPO} synthesis procedure \myalgo\ uses, so the two approaches are directly comparable. Starting from the center of the search space, BayesOpt iteratively proposes 3 candidates per round for 20 rounds ($\approx\!60$ evaluations per template-target pair, $\approx\!200$ evaluations in aggregate across templates per target). We sweep three performance targets $\mathrm{J}_\text{target}\!\in\!\{200,300,450\}$ (the oracle attains $456.2\pm4.6$), and for each plot the reciprocal resource cost of the best feasible parameterization found so far against the number of evaluations.

\paragraph{Findings.} \texttt{BO-MSC}'s success is highly contingent on both the chosen template and the performance target (\Cref{fig:p2d_result_BOpt}). At the loose target $\mathrm{J}_\text{target}\!=\!200$, only \texttt{T1} eventually reaches \myalgo's resource cost, and only after $\approx\!48$ evaluations. At the stringent near-oracle target $\mathrm{J}_\text{target}\!=\!450$, the finely-banded \texttt{T3} matches \myalgo\ and \texttt{T4} marginally exceeds it, but only after $\approx\!30$ evaluations. For the intermediate target $\mathrm{J}_\text{target}\!=\!300$, no template reaches \myalgo's frontier within the budget. In short, even when \texttt{BO-MSC} edges out \myalgo, it does so only with (a) a pre-specified template that already encodes the right structural prior and (b) runs more evaluations (as it is uninformed + most suitable template is unknown a-priori). Whereas \myalgo\ discovers comparable or better designs from scratch by jointly searching over structure and parameters. Combining the complementary strengths of LLM-guided structural search with BayesOpt for fine continuous-parameter tuning is a promising direction for future work.

\subsection{Alternative LLM-in-the-loop mechanisms}
\label{app:compare:llm}
\myalgo\ wraps the LLM in a genetic algorithm with a population, LLM crossover, feasible-first tournament selection, and a compacted \texttt{Design Log} that summarizes past evaluations as cross-iteration memory (\Cref{app:designer:loop}). To assess which of these ingredients actually drive the search, we ablate the loop into five variants outlined in \Cref{tab:llm:variants} for the \task{clutter-nav} task. These variants isolate (i) the genetic algorithm machinery and (ii) the design log. All variants are otherwise identical -- same task, seed designs, MPPI controller synthesis, and LLM (\stt{gemini-3.1-pro}) -- so any difference is attributable to the search harness alone.

\begin{table}[H]
  \centering\small
  {\renewcommand{\arraystretch}{1.2}
  \begin{tabular}{llcc}
    \toprule
    {\bf Variant (legend)} & {\bf Search mechanism} & {\bf Pop. + Crossover} & {\bf Cross-Iter.\ Memory} \\
    \midrule
    \textsf{GA + Design Log} (\myalgo) & feasible-first GA            & \checkmark & design log \\
    \textsf{GA}                        & feasible-first GA           & \checkmark & none \\
    \textsf{Refine Best + Design Log}  & greedy hill-climb (best)    & --         & design log \\
    \textsf{Refine Latest + Design Log}& greedy hill-climb (latest)  & --         & design log \\
    \textsf{Single Chat}               & one growing conversation    & --         & raw chat history \\[1em]
  \end{tabular}}
  \caption{The five LLM-in-the-loop variants compared in \Cref{fig:LLM-ablation}. \ssf{GA} removes only the design log from \myalgo; the two \ssf{Refine} variants replace the population/crossover with greedy single-design refinement (of the global best or the most recent design, respectively) while keeping the design log; \ssf{Single Chat} additionally drops the design log, letting an unbounded conversation history serve as the only memory.}
  \label{tab:llm:variants}
\end{table}

\begin{figure}[H]
    \centering
    \includegraphics[width=\linewidth]{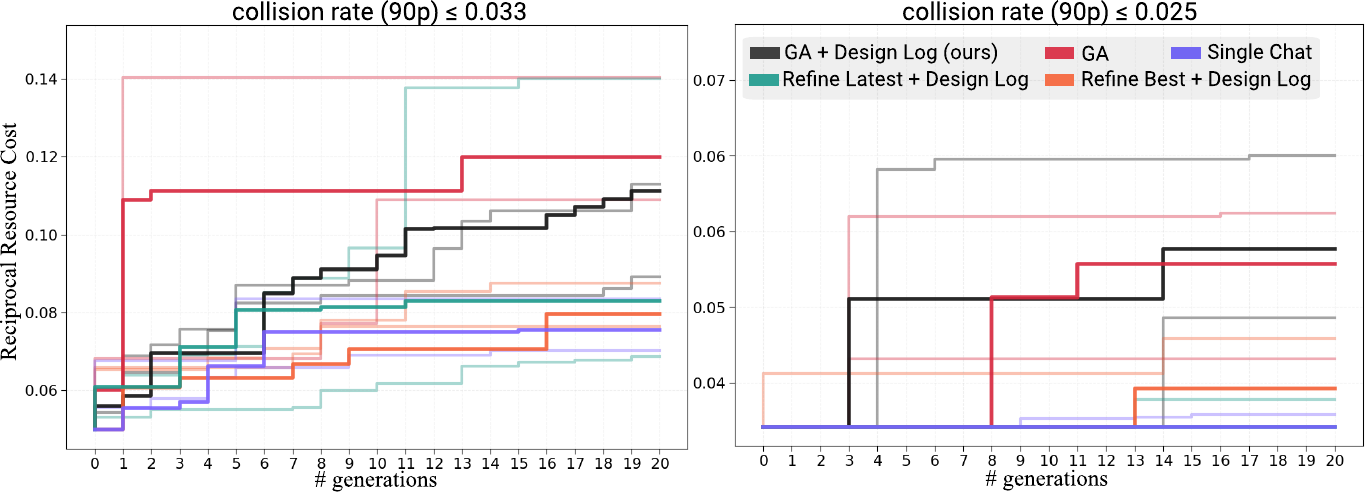}
    \caption{Ablation of \myalgo's LLM-in-the-loop search mechanism on \task{clutter-nav} (\Cref{tab:llm:variants}). Each panel plots the reciprocal resource cost of the best feasible design found so far against the generation count for a reciprocal collision-rate target ($1/\mathrm{cpt}_{90}\!=\!30$, i.e.\ $\mathrm{cpt}_{90}\!\le\!0.033$, left; and $40$, i.e.\ $\le\!0.025$, right) under the $\ge\!80\%$ success constraint; the median of 3 seeds is represented in bold and individual seeds are marked in faint colors.
    Both GA variants (\ssf{GA + Design Log}, \ssf{GA}) reach substantially cheaper designs than the greedy single-design (\ssf{Refine Best}/\ssf{Latest}) and \ssf{Single Chat} variants -- the GA harness clearly drives improvement, whereas adding the design log to the GA yields no clear advantage.}
    \label{fig:LLM-ablation}
\end{figure}

\paragraph{Setup.} We run the ablation on \task{clutter-nav} (\Cref{app:nav:setup}), under the success constraint $\ge\!80\%$ and two reciprocal collision-rate targets $1/\mathrm{cpt}_{90}\!\in\!\{30,40\}$ (i.e.\ $\mathrm{cpt}_{90}\!\le\!\{0.033,0.025\}$). Each variant is run for 3 seeds and up to 20 generations, and we track the reciprocal resource cost of the best feasible design found so far against the generation count (\Cref{fig:LLM-ablation}; median seed in bold, individual seeds in faint colors).

\paragraph{Findings.} Two trends emerge in \Cref{fig:LLM-ablation}. First, the GA harness is the dominant factor: both GA variants -- with \emph{and} without the design log -- converge to cheaper feasible designs compared to the three single-design variants, which plateau $\approx\!50$--$60\%$ higher in resource cost. The population and LLM crossover inject the stochastic diversity needed to escape the local optima that greedy hill-climbing (\ssf{Refine Best}/\ssf{Refine Latest}) and the single growing conversation (\ssf{Single Chat}) settle into early. Second, the effect of the design log is equivocal: removing it from the GA does not hurt -- that variant often reaches low-cost designs at least as quickly -- whereas among the greedy variants the design log drives steady refinement but can also anchor the search to early suboptimal commitments. We therefore conclude that the GA framework is essential, while the design log's contribution is setting-dependent and inconclusive.

\section{Extended Related Work}
\label{app:related_work}
\paragraph{System Co-Design.}
\citet{censi2015mathematical} presents monotone co-design theory which provides a tool to compositionally reason about resource-functionality trade-offs in system design. While this presents a useful formalism, several components of the design space need to be extensively characterized analytically or empirically~\citep{neumann2024co}. \myalgo\ can be viewed as a complementary tool in the larger co-design problem to help empirically characterize resource-performance trade-offs in controller synthesis.

\paragraph{Minimalist Robots.} A classical line of work characterizes a notion of \emph{least} information a robot needs to solve a task or represent its controller, such as: action-based sensors~\citep{erdmann1995understanding,donald1995information}, sensor lattices~\citep{lavalle2012sensing,okane2008comparing}, motion description languages~\citep{egerstedt1999motion}. These results give lower bounds and analytic constructions of minimal information sensor/controller specification for specific tasks. \myalgo\ hinges on similar notions, but relies on the LLM's potential as a useful idea generator to scalably explore the vast design space and provide constructive evidence of resource-efficient designs on more complex tasks.

\end{document}